\documentclass[10pt,twocolumn,letterpaper,usenames,dvipsnames]{article}

\usepackage{iccv}
\usepackage{times}
\usepackage{epsfig}
\usepackage{graphicx}
\usepackage{float}
\usepackage{amsmath}
\usepackage{amssymb}
\usepackage{soul}
\usepackage{multicol}
\usepackage{blindtext}
\usepackage{here}
\usepackage{caption}
\usepackage{makecell}
\usepackage[normalem]{ulem}
\usepackage{multirow}
\usepackage{booktabs}
\usepackage{xcolor, colortbl}
\usepackage{xspace}
\usepackage{balance}
\definecolor{Gray}{gray}{0.9}
\usepackage[pagebackref=true,breaklinks=true,letterpaper=true,colorlinks,bookmarks=false]{hyperref}

\newcommand{\gtShape}{\mathbf{\beta}}
\newcommand{\gtPose}{\mathbf{\theta}}

\newcommand{\methodname}{PARE\xspace} 

\newcommand{\figref}[1]{Fig.~\ref{#1}}

\newcommand{\mpi}{\texttt{MPI-INF-3DHP}\xspace}
\newcommand{\mpii}{\texttt{MPII}\xspace}
\newcommand{\lspet}{\texttt{LSPET}\xspace}
\newcommand{\hthreesixm}{\texttt{Human3.6M}\xspace}
\newcommand{\threedpw}{\texttt{3DPW}\xspace}
\newcommand{\threedpwocc}{\texttt{3DPW-OCC}\xspace}
\newcommand{\coco}{\texttt{COCO}\xspace}
\newcommand{\cocoeft}{\texttt{COCO-EFT}\xspace}
\newcommand{\ooh}{\texttt{3DOH}\xspace}

\newcommand{\supmat}{Sup.~Mat.\xspace}

\newcommand{\real}{\mathbb{R}}

\iccvfinalcopy 
\renewcommand{\etal}{et al.\xspace}
\renewcommand{\ie}{i.e.\xspace}
\renewcommand{\eg}{e.g.\xspace}

\def\iccvPaperID{1275} 

\ificcvfinal\fi

\begin{document}
	
	\title{{\methodname}: Part Attention Regressor for 3D Human Body Estimation}
	\author{%
		Muhammed Kocabas$^{1,2}$\quad \; Chun-Hao P. Huang$^1$\quad \; Otmar Hilliges$^2$\quad \; Michael J. Black$^1$\\\
		\normalsize $^1$Max Planck Institute for Intelligent Systems, T\"{u}bingen, Germany \quad
		\normalsize $^2$ETH Zurich\\
		\normalsize \texttt{\{\href{mailto:mkocabas@tue.mpg.de}{mkocabas},\href{mailto:nathanasiou@tue.mpg.de}{paul.huang},\href{mailto:black@tue.mpg.de}{black}\}@tue.mpg.de} \quad 
		\texttt{\href{mailto:otmar.hilliges@inf.ethz.ch}{otmar.hilliges@inf.ethz.ch} }
	}
	
	\twocolumn[{%
		\renewcommand\twocolumn[1][]{#1}%
		\maketitle
		\begin{center}
			\newcommand{\teaserwidth}{\textwidth}
			\vspace{-0.15in}
			\centerline{
				\includegraphics[width=0.95\teaserwidth,clip]{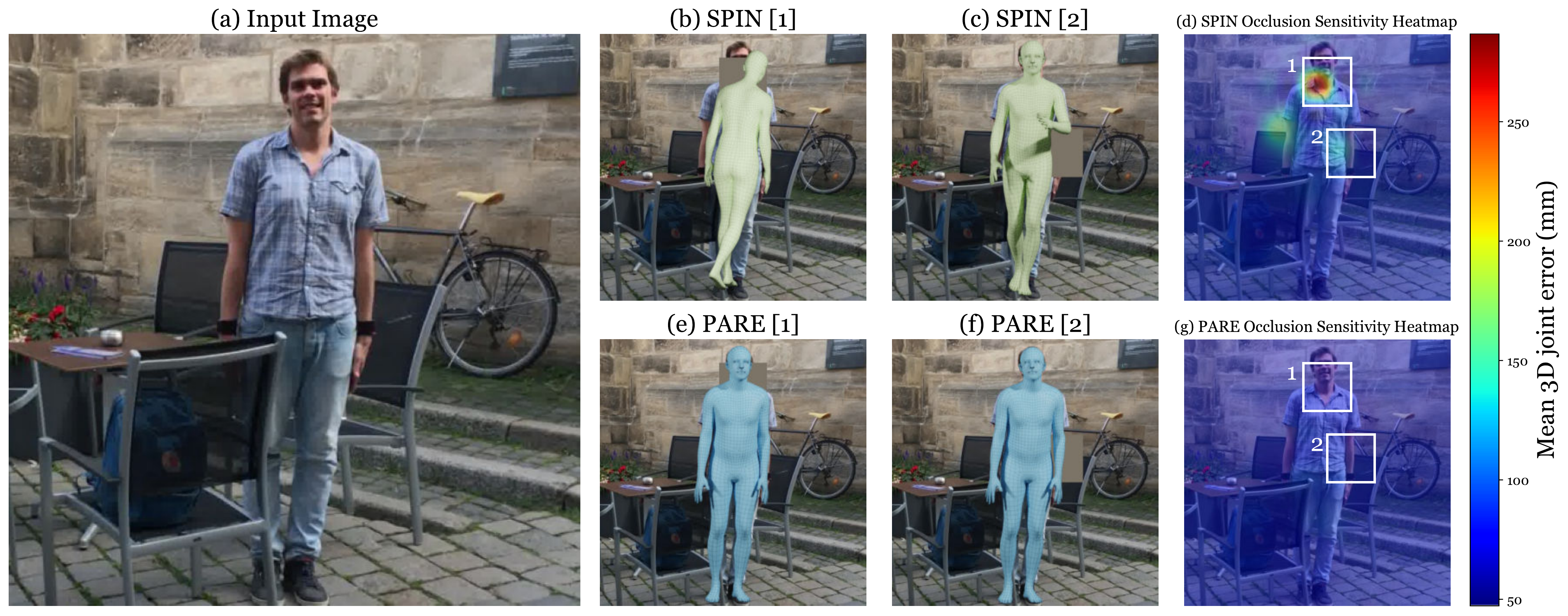}
			}
			\vspace{-0.1in}
			\captionof{figure}{{\bf Occlusion Sensitivity Analysis.} Given an input image (a), a small occluding patch (shown in gray) causes SPIN~\cite{SPIN:ICCV:2019} to fail (b,c), whereas our method (\methodname) (e,f) is robust to the occluder. Sub-figures on the right show the sensitivity of SPIN (d) and PARE (g) to an occluding patch (the size of the white squares) centered at every point in the image. Warmer colors mean higher average joint error.  
			}
			\vspace{-0.05in}
			\label{fig:teaser}
		\end{center}%
	}]
	
\begin{abstract}
	
	
	Despite significant progress, we show that state of the art 3D human pose and shape estimation methods remain sensitive to partial occlusion and can produce dramatically wrong predictions although much of the body is observable.  
	To address this, we introduce a soft attention mechanism, called the \emph{Part Attention REgressor ({\methodname})}, that learns to predict body-part-guided attention masks. 
	We observe that state-of-the-art methods 
	rely on global feature representations, making them sensitive to even small occlusions. 
	In contrast, \methodname's  part-guided attention mechanism overcomes these issues by 
	exploiting information about the visibility of individual body parts while leveraging information from neighboring body-parts to predict occluded parts. 
	We show qualitatively that {\methodname} learns sensible attention masks, and quantitative evaluation confirms that {\methodname} achieves more accurate and robust reconstruction results than existing approaches on both occlusion-specific and standard benchmarks.  The code and data are available for research purposes at {\small \url{https://pare.is.tue.mpg.de/}}
	\vspace{-2ex}

\end{abstract}
	\section{Introduction}
\label{introduction}
Regressing 3D human pose and shape (HPS) directly from RGB images has many applications in robotics, computer graphics, AR/VR and beyond.
The task is to take a single image \cite{kanazawa_hmr, SPIN:ICCV:2019,pavlakos2018humanshape} or video sequence  \cite{kanazawa_temporal_hmr,kocabas2019vibe,Luo2020MEVA} as input and to regress the parameters of a human body model such as SMPL \cite{looper_smpl} as output. 
Powered by deep CNNs, this task has seen rapid progress  \cite{kanazawa_hmr, kocabas2019vibe,SPIN:ICCV:2019,pavlakos2018humanshape}. 
However, in fully in-the-wild settings, people often appear under occlusion either due to self-overlapping body-parts, due to close-range interaction with other people or due to occluding objects such as furniture or other scene content. 
While pose estimation under occlusion has been treated in the literature \cite{cheng20203d,cheng2019occlusion,ghiasi2014parsing,huang2009estimating,rafi2015semantic,Rockwell2020,vosoughi2018deep,wang20203d,zhangoohcvpr20}, we highlight that this issue is particularly important in the context of direct regression methods. Such methods use all the pixels in the input to predict a single set of pose and shape parameters. Thus their pose estimates are particularly sensitive to even small perturbations in the observations of the body and its parts. 

In this paper, we apply a visualization technique \cite{zeiler2014visualizing} for occlusion sensitivity analysis that
yields insights into when and why such methods fail. 
This indicates that, for state-of-the-art (SOTA) methods, relatively small occlusions, even of only a single joint, can lead to entirely implausible pose predictions. 
This is illustrated in \figref{fig:teaser}, where we slide an occluder over the image, regress body pose, and compute the average 3D joint error with respect to ground truth.
The heatmaps in \figref{fig:teaser} (d,g) illustrate a method's sensitivity to a square occluder centered at each pixel location (shown in white). 
The visualization reveals that methods like SPIN \cite{pavlakos2018humanshape} are highly sensitive to localized part occlusion.
To address this issue, we propose a method, based on a novel part-guided attention mechanism, making direct regression approaches more robust to occlusion.

The proposed method is called Part Attention REgressor (\methodname). It has two tasks: the primary one is learning to regress 3D body parameters in an end-to-end fashion, and the auxiliary task is learning attention weights per body part. Each task has its own pixel-aligned feature extraction branch. We guide the attention branch with part segmentation labels in the early stages of training and continue without them for the later stages, thus we call it \emph{body-part-driven attention}.
Our key insight is that, to be robust to occlusions, the network should leverage pixel-aligned image features of visible parts to reason about occluded parts.

Given the success of attention-based methods on other tasks \cite{expose2020eccv,epipolartransformers2020cvpr,Lu_2019_CVPR,Wang_2018_CVPR}, we exploit insights gained from the occlusion sensitivity analysis to focus attention on body parts. Therefore, we supervise the attention mask with part segmentations, but then train end-to-end with pose supervision only, allowing the attention mechanism to leverage all useful information from the body and the surrounding pixels. This gives the network freedom to attend to regions it finds informative in an unsupervised way. %
As a result, {\methodname} learns to rely on visible parts of the body to improve robustness to occluded parts and overall performance on 3D pose estimation (\figref{fig:teaser} e-f).

To quantitatively evaluate the performance of {\methodname}, we perform experiments on the 3DPW \cite{vonMarcard2018_3dpw}, 3DOH \cite{zhangoohcvpr20}, and 3DPW-OCC \cite{vonMarcard2018_3dpw} datasets.
The results show that {\methodname} yields consistently lower error than the state-of-the-art for both occlusion and non-occlusion cases.

In  summary,  our  key  contributions   are:
(1) We apply a visualization technique \cite{zeiler2014visualizing} to study how local part occlusion can influence global pose; we call this occlusion sensitivity analysis.
(2) This analysis motivates a novel body-part-driven attention framework for 3D HPS regression that leverages pixel-aligned localized features to regress body pose and shape. 
(3) The network uses part visibility cues to reason about occluded joints by aggregating features from the attended regions, and by doing so, achieves robustness to occlusions. 
(4) We achieve SOTA results on a 3D pose estimation benchmark featuring occluded bodies, as well as a standard benchmark.
	\section{Related Work}
\label{related_work}
We focus on 3D human shape and pose estimation from RGB images and discuss how previous approaches handle occlusions in various scenarios, \eg self occlusion, camera frame occlusion, and scene object occlusion.

{\bf 3D pose and shape from a single image.} 
In estimating human shape and pose, many methods output the parameters of 3D human body models~\cite{scape,looper_smpl,SMPL-X:2019}.
Initial work predicts the 3D body using 
keypoints and silhouettes~\cite{agarwal2006recovering,balan2008,Balan:CVPR:2007,grauman2003inferring,sigal2008combined}.
These approaches are fragile, need manual input, use additional data, \eg~multi-view images, or do not generalize well to in-the-wild images.
SMPLify \cite{bogo_smplify} was the first automated method to fit the SMPL model to the output of a 2D keypoint detector~\cite{Leonid2016DeepCut}. Lassner~\etal~\cite{lassner_up3d} employ silhouettes together with keypoints during fitting. 
In contrast, deep neural networks regress SMPL parameters directly from pixels~\cite{guler_2019_CVPR,kanazawa_hmr,omran2018nbf,pavlakos2018humanshape,Tan,tung2017self}. 
In order to deal with the lack of in-the-wild 3D ground-truth, methods use a 2D keypoint re-projection loss as weak supervision~\cite{kanazawa_hmr,Tan,tung2017self}, use intermediate 2D representations, \eg~body/part segmentation~\cite{omran2018nbf,pavlakos2018humanshape,zanfir2020weakly}, 2D sparse keypoints~\cite{sengupta2020synthetic,zanfir2020weakly}, or leverage a human in the loop \cite{lassner_up3d}.
Note that the use of part segmentation in \cite{lassner_up3d,omran2018nbf,zanfir2020weakly} is very different from our approach, in which part segmentations are used to facilitate soft attention.
Kolotouros \etal~\cite{SPIN:ICCV:2019} combine HMR~\cite{kanazawa_hmr} and SMPLify~\cite{bogo_smplify} in a training loop. At each step, HMR initializes SMPLify, which fits the body model to 2D joints, resulting in better supervision for the network. 
The above methods are typically sensitive to occlusion.

\begin{figure*}[t]
	\centering
	\includegraphics[width=\textwidth]{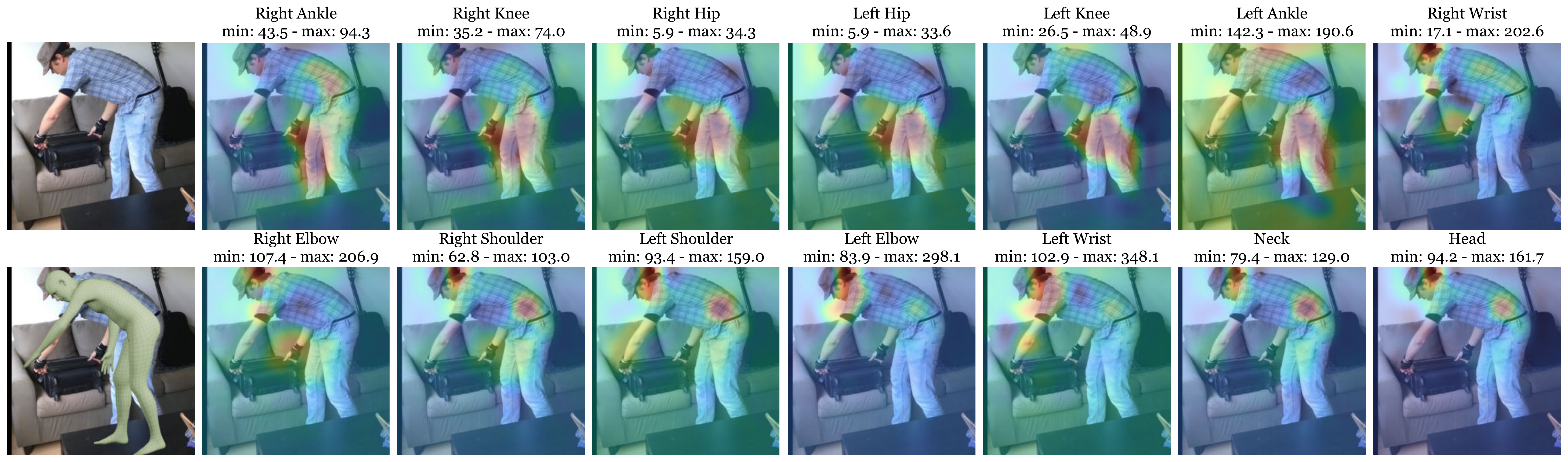}
	\vspace{-0.25in}
	\caption{{\bf Occlusion sensitivity analysis.} Heatmaps illustrate the error of SPIN \cite{SPIN:ICCV:2019} in individual joints caused by an occluder placed at each image location.
		Image size: $224\times224$; occluding patch: $40\times40$. The title of each heatmap names the joint and notes the range of the 3D error in mm visualized in the heatmap.
		See Section~\ref{sec:occlusion_analysis} for analysis.
	}
	\label{fig:occlusion_analysis}
\end{figure*}{}

{\bf Implicit occlusion handling (data augmentation).}
Ideally, the regressed 3D body should be the same with or without occlusion.
Current SOTA pose and shape estimation methods~\cite{kanazawa_hmr,kocabas2019vibe,SPIN:ICCV:2019} directly encode the entire input region as one CNN feature after global average pooling, followed by body model parameter regression. 
The lack of pixel-aligned structure makes it hard for networks to explicitly reason about the locations and  visibility of body parts. 
A common way to achieve robustness to occlusion in these frameworks is through data augmentation. 
For example, frame occlusion is often simulated by cropping~\cite{biggs2020multibodies,joo2020eft,Rockwell2020}, whereas object occlusion is approximated by overlaying object patches on the image~\cite{georgakis2020hierarchical,sarandi2018robust}. Instead of applying augmentation to input images, Cheng~\etal~\cite{cheng20203d} apply augmentations to heatmaps that contain richer semantic information and hence occlusions can be simulated in a more intelligent way.
While helpful, these synthetic occlusions do not fully capture the complexity of occlusions in realistic images, nor do they provide insight into how to improve the network architecture to be inherently more robust to occlusion.

{\bf Explicit occlusion handling.}
To reason more explicitly about occlusions, previous work exploits visibility information. 
For example, Cheng~\etal~\cite{cheng2019occlusion} avoid including occluded joints when computing losses during training. Such visibility information is obtained by approximating the human body as a set of cylinders, which is not realistic and only handles self occlusion. Wang~\etal~\cite{wang20203d} learn to predict occlusion labels to zero out occluded keypoints before applying temporal convolution over a sequence of 2D keypoints.

Person-person occlusion is particularly common and challenging. 
For multi-person regression, Jiang \etal~\cite{jiang2020mpshape} use an interpenetration loss to avoid collision and an ordinal loss to resolve depth ambiguity. 
Sun \etal~\cite{CenterHMR} estimate all people in an image simultaneously, enabling their method to learn about person-person occlusion.
While \cite{CenterHMR} learns features that are robust to person-person occlusion, {\methodname} learns to focus attention on individual body parts.

Zhang \etal~\cite{zhangoohcvpr20} leverage saliency masks as visibility information to gain robustness to scene/object occlusions. 
Human meshes are parameterized by UV maps where each pixel stores the 3D location of a vertex, and occlusions are cast as an image-inpainting problem.
The requirement of accurate saliency maps limits the performance on in-the-wild images. Furthermore, UV-coordinates can result in mesh artifacts, as shown in \supmat

	\section{Occlusion Sensitivity Analysis}
\label{sec:occlusion_analysis}
To extract features from the input image region $I$, current direct regression approaches~\cite{kanazawa_hmr,SPIN:ICCV:2019} use a ResNet-50~\cite{he2016resnet} backbone and take the features after global average pooling (GAP), followed by an MLP that regresses and refines the parameters iteratively. 
In this section, we investigate the impact of occlusions on this type of architecture. Our analysis is inspired by  Zeiler \etal~\cite{zeiler2014visualizing} who systematically cover different portions of the image with a gray square to analyze how feature maps and classifier output changes. 
In contrast, we slide a gray occlusion patch over the image 
and regress body poses using SPIN~\cite{SPIN:ICCV:2019}.
Instead of computing a classification score as in \cite{zeiler2014visualizing}, we measure the per-joint Euclidean distance between ground truth and predicted joints. 
We create an error heatmap, in which each pixel indicates how much error the model creates for joint $j$ when the occluder is centered on this pixel.
In addition to per-joint heatmaps, we compute an aggregate occlusion sensitivity map, that shows how the average joint error is influenced by an occlusion; this is visualized in \figref{fig:teaser}(d) and in greater detail in the \supmat

\begin{figure}
	\centering
	\includegraphics[width=\columnwidth]{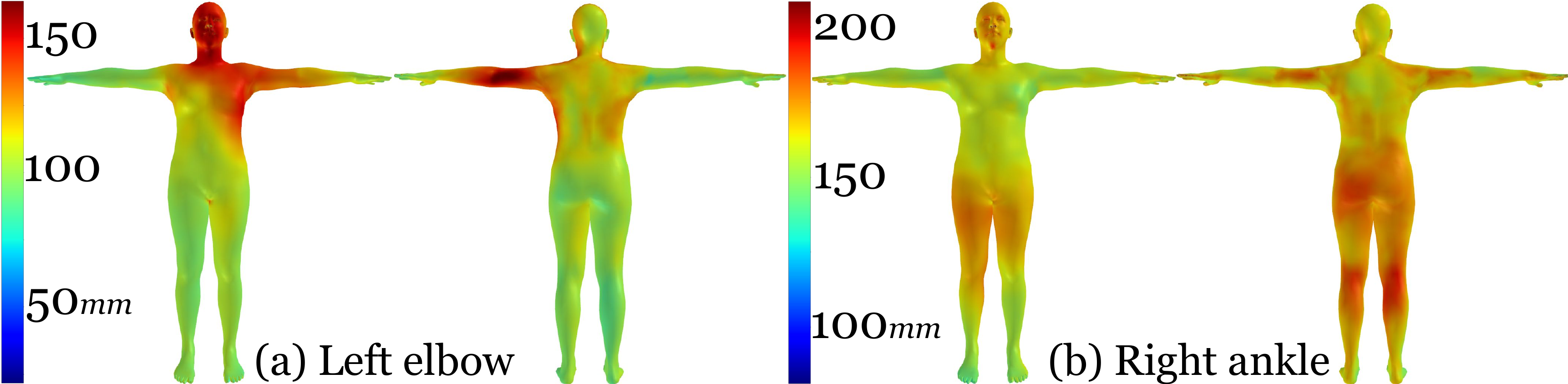}
	\caption{Occlusion sensitivity meshes for SPIN~\cite{SPIN:ICCV:2019}. 
	}
	\label{fig:occ_meshes}
	\vspace{-2ex}
\end{figure}{}

\begin{figure*}[h]
	\centerline{
		\includegraphics[width=\textwidth]{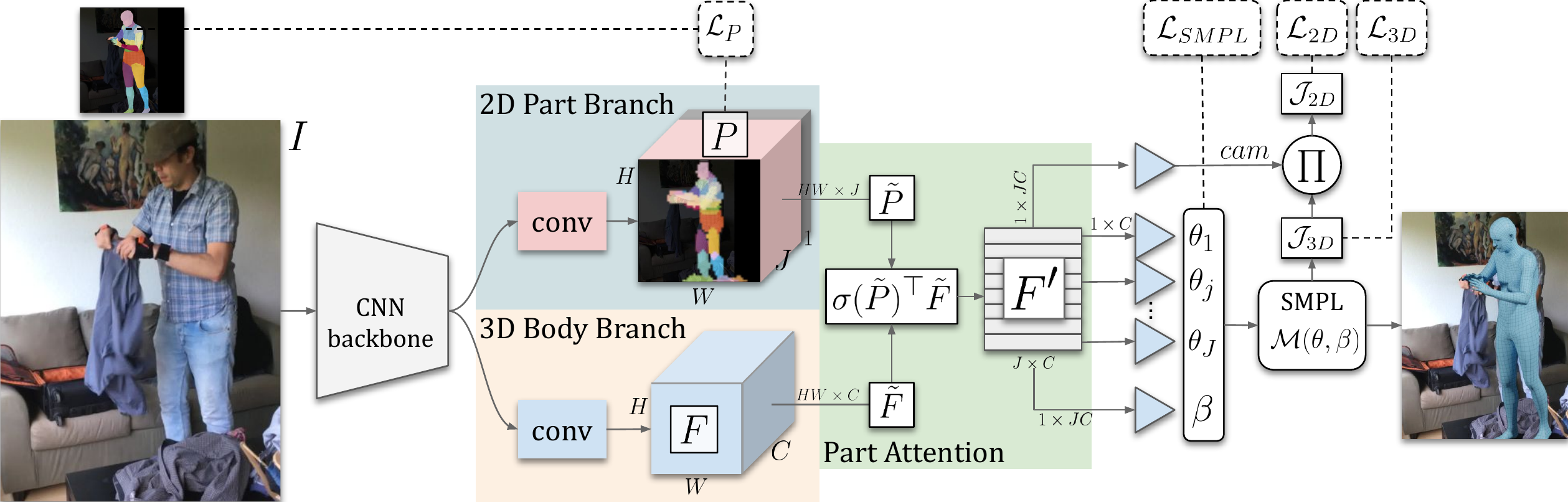}}
	\caption{\textbf{{\methodname} model architecture.} 
		Given an input image, {\methodname} extracts two pixel-level features $P$ and $F$, which are fused by part attention (green box) leading to the final feature $F^\prime$ for camera and SMPL body regression.
	}
	\label{fig:model}
	\vspace{-2ex}
\end{figure*}{}

The per-joint error heatmaps for SPIN are visualized in \figref{fig:occlusion_analysis} for a sample image from the 3DPW dataset~\cite{vonMarcard2018_3dpw}.
Each sub-image corresponds to a particular joint and hot regions are locations where occlusion causes high error in this joint.  
This visualization allows us to make several observations.
(1) Errors are low in the background and high on the body. This shows that SPIN has learned to attend to meaningful regions.
(2) Joints visible in the original image have high errors when they are occluded by the square, as expected.
(3) For joints that are naturally occluded, the network relies on other regions to reason about the occluded poses. 
For example, in the top row of Fig.~\ref{fig:occlusion_analysis}, we observe high errors for the left/right ankles (which are occluded) when we occlude the thigh region.
Since the network has no image features for the occluded parts, it must look elsewhere in the image for evidence.
(4) Such dependencies happen not only between neighboring parts; occlusion can have long-range effects (e.g.~occluding the pelvis causes errors in the head). 

We further overlay the estimated body on the heatmap to transfer the per-pixel error to visible vertices.
We run this analysis over the complete 3DPW dataset, pool the per-vertex error across the dataset and visualize the result on a SMPL body model, giving one \emph{occlusion sensitivity mesh} per joint.
For example, Fig.~\ref{fig:occ_meshes}(a) shows that the left elbow is sensitive to occlusion of the face, the left shoulder and the left upper arm region. See \supmat~for more examples. 
\section{Method}
\label{methods}

Given the observations above, {\methodname} is designed with the following insights.
First, as shown in Fig.~\ref{fig:occlusion_analysis}, SOTA networks~\cite{kanazawa_hmr,kocabas2019vibe,SPIN:ICCV:2019} learn to attend to meaningful regions implicitly, despite limited spatial information after global average pooling.
To better understand whether body parts are visible or not, and to know if their locations are occluded, {\methodname} exploits a pixel-aligned structure, where each pixel corresponds to a region in the image and stores a pixel-level representation, namely, a feature volume.
Second, since estimating attention weights and learning end-to-end trainable features for 3D poses are two different tasks, {\methodname} is equipped with two feature volumes: one from the 2D part branch that estimates attention weights and one from the 3D body branch that performs SMPL parameter regression.
Finally, to model the body part dependencies observed above, {\methodname} exploits part segmentations as soft attention masks to adjust the contribution of each feature in the 3D body branch differently for each joint. 

\textbf{Preliminaries: Body Model.}
\label{sec:body_model}
SMPL~\cite{looper_smpl} represents the body pose and shape by $\Theta$, which consists of the pose $\gtPose \in \real^{72}$ and shape $\gtShape \in \real^{10}$ parameters. 
Here we use the gender-neutral shape model as in previous work \cite{kanazawa_hmr,SPIN:ICCV:2019}.
Given these parameters, the SMPL model is a differentiable function that outputs a posed 3D mesh $\mathcal{M}(\theta,\beta) \in \real^{6890 \times 3}$.
The 3D joint locations $\mathcal{J}_\mathit{3D}=W \mathcal{M} \in \real^{J\times3}$, $J=24$, are computed with a pretrained linear regressor $W$.

\subsection{Model Architecture and Losses}
\label{sec:model}
The overall framework of {\methodname} is depicted in Fig.~\ref{fig:model}. 
Our architecture works as follows: given an image $I$, we first run a CNN backbone to extract \emph{volumetric} features, \eg~before the global average pooling layer
for ResNet-50, followed by two separate feature extraction branches to obtain two volumetric image features. 
We denote the 2D part branch as $P \in \mathbb{R}^{H\times W \times (J+1)}$, modelling $J$ part attention and $1$ background masks, where $H$ and $W$ are the height and width of the feature volume and each pixel $(h,w)$ stores the likelihood 
of belonging to a body part $j$. 
The other branch, denoted by $F \in \mathbb{R}^{H\times W \times C}$, is used for 3D body parameter estimation.
It has the same spatial dimensions $H\times W$ as $P$ but a different number of channels, $C$.

Let $P_j \in \mathbb{R}^{H\times W}$ and $F_c \in \mathbb{R}^{H\times W}$ denote the $j$-th and $c$-th channel of $P$ and $F$, respectively, and let $F^\prime \in \mathbb{R}^{J\times C}$ represent the final feature tensor. 
Each element in $F_c$ contributes proportionally to $F^\prime$ according to the corresponding elements in $P_j$ after spatial softmax normalization $\sigma$. 
Formally, the element at location $(j,c)$ in $F^\prime$ is computed as:
\begin{equation}
\label{eq-softattention}
F^\prime_{j,c}= \sum_{h,w} \sigma(P_j) \odot F_c,
\end{equation}
where $\odot$ is the Hadamard product. 
In other words, we use $\sigma(P_j)$ as a soft attention mask to aggregate features in $F_c$.
This operation can be efficiently implemented as a dot product similar to existing attention implementations: $F^{\prime} = \sigma( \tilde{P})^{\top}\tilde{F}$, where $\tilde{P} \in \mathbb{R}^{HW \times J}$ and $\tilde{F} \in \mathbb{R}^{HW \times C}$ denote the reshaped $P$ (omitting the background mask) and $F$ respectively. This attention operation suggests that if a particular pixel has a higher attention weight, its corresponding feature contributes more to the final representation $F^{\prime}$. 
We supervise the 2D part branch $P$ with ground-truth segmentation labels, which helps the attention maps of \emph{visible} parts converge to the corresponding regions.
For \emph{occluded} parts, however, this encourages $0$ attention weights for all pixels in $P_j$ because they do not exist in the ground-truth segmentation labels. 
An attention map with all $0$ weights is undesirable and, in practice, also impossible since the spatial softmax ensures that all elements sum to $1$.
Therefore, we adopt a hybrid approach that supervises the 2D part branch only for the initial stage and continues training without any supervision. This allows the network to attend to other regions to estimate the poses of an occluded joint.

We take the full feature tensor $ F^\prime$ to regress body shape $\beta$ and a weak-perspective camera model with scale and translation parameters $[s,t], t \in \real^2$, while each row, $ F^\prime_j$, is also sent to different MLPs to predict the rotation of each part, $\theta_j$, parameterized as a 6D vector following~\cite{kocabas2019vibe,SPIN:ICCV:2019} \footnote{With slight abuse of notations, $\theta$ is in axis-angle form when passed to the SMPL model but in 6D-vector form during the regression and loss computation.}.

Overall, our total loss is: 
\begin{equation}
\mathcal{L} = \lambda_{3D}\mathcal{L}_{3D} + \lambda_{2D}\mathcal{L}_{2D} + \lambda_{SMPL} \mathcal{L}_{SMPL} + \lambda_P \mathcal{L}_{P},
\end{equation}
where each term is calculated as:
\begin{align*}
\mathcal{L}_{\mathit{3D}} &=   \| \mathcal{J}_{\mathit{3D}} \; - \; \hat{\mathcal{J}}_{\mathit{3D}} \|_F^2 , \\
\mathcal{L}_{\mathit{2D}} &=  \| \mathcal{J}_\mathit{2D} \; - \; \hat{\mathcal{J}}_\mathit{2D} \|_F^2 , \\
\mathcal{L}_{\mathit{SMPL}} &= \| \Theta \; - \; \hat{\Theta} \|_2^2, \\
\mathcal{L}_P &= \frac{1}{HW} \sum_{h,w} \text{CrossEntropy}\left(\sigma(P_{h,w}), \hat{P}_{h,w}\right),
\end{align*}
where $\hat{x}$ represents the ground truth for the corresponding variable $x$. 
To compute the 2D keypoint loss, we need the SMPL 3D joint locations $\mathcal{J}_\mathit{3D}(\theta, \beta) = W \mathcal{M}(\theta, \beta)$, which are computed from the body vertices with a pretrained linear regressor $W$.
With the inferred weak-perspective camera, we compute the 2D projection of the 3D joints $\mathcal{J}_\mathit{3D}$, as
$\mathcal{J}_\mathit{2D} \in \mathbb{R}^{J \times 2} = s\Pi(R\mathcal{J}_\mathit{3D}) + t $, where $R \in SO(3)$ is the camera rotation matrix and $\Pi$ is the orthographic projection. $\lambda$ is a scalar coefficient to balance the loss terms. 
Let $P_{h,w} \in \mathbb{R}^{1\times1 \times (J+1)}$ denote the fiber
of $P$ at the location $(h,w)$, 
and $\hat{P}_{h,w} \in \{0,1\}^{(J+1)}$ denotes the ground-truth part label at the same location, expressed as a one-hot vector.
The part segmentation loss $ \mathcal{L}_P$ is the cross-entropy loss between $P_{h,w}$ after softmax and $\hat{P}_{h,w}$, averaged over $H\times W$ elements. 
Note that this softmax normalizes along the fiber $P_{h,w}$ while the one in Eq.~\ref{eq-softattention} normalizes over the slice $P_j$.

\subsection{Implementation Details}
\label{sec:implementation}
As mentioned above, the body-part label supervision via $\mathcal{L}_P$ is applied on the attention tensor $P$ only in the \emph{initial} stages of training. 
It is later removed by setting $\lambda_P$ to zero, turning the attention mechanism into an unsupervised pure soft-attention. 
The absence of body-parts due to occlusion is the main motivation for this training scheme. 
Setting $\lambda_P$ to zero allows the attention mechanism to also consider pixels beyond the body itself. 
Hence, the final attention maps do not necessarily (and often do not) resemble body part segmentations, as shown later in Fig.~\ref{fig:attention} and \supmat
If a body part is visible, it focuses on that part directly; if it is occluded, the attention is free to leverage other informative regions in the image. 
In Sec.~\ref{experiments}, we analyze how the accuracy of part segmentation impacts body reconstruction.

We evaluate both ResNet-50 \cite{he2016resnet} and HRNet-W32 \cite{hrnet} networks as the backbone. 
Since ResNet-50 is widely used in other SOTA methods~\cite{kanazawa_hmr,kocabas2019vibe,SPIN:ICCV:2019}, we choose it as the default backbone for most of the experiments unless stated otherwise. 
We extract the $7\times7\times2048$ feature volumes before global average pooling. For the 2D and 3D branches, we use three $2\times$ upsampling followed by $3\times3$ convolutional layers applied with batch-norm and ReLU. The number of conv kernels is 256. 
For HRNet-W32, since it already provides volumetric features with a higher resolution, we only use two $3\times3$ convolutional layers applied with batch-norm and ReLU as the 2D and 3D branches. 

To obtain part attention maps, we apply $J+1$ $1\times1$ convolutional kernels to 2D part features to reduce the channel dimension. 
After obtaining the $J\times C$ final feature $F^\prime$, we use separate linear layers to predict each SMPL joint rotation $\theta_j$. We regress shape and camera parameters from the flattened $F^\prime$ vector. We use a fixed image size of $224\times224$ for all experiments. The Adam optimizer with a learning rate of $5\times10^{-5}$ and batch size 64 is used to optimize our model. PARE is end-to-end trainable in a single stage, unlike recent multi-stage methods~\cite{Choi_2020_ECCV_Pose2Mesh,guler_2019_CVPR,Moon_2020_ECCV_I2L-MeshNet,zanfir2020weakly}.
	\begin{figure}[t]
	\centering
	\includegraphics[width=\columnwidth]{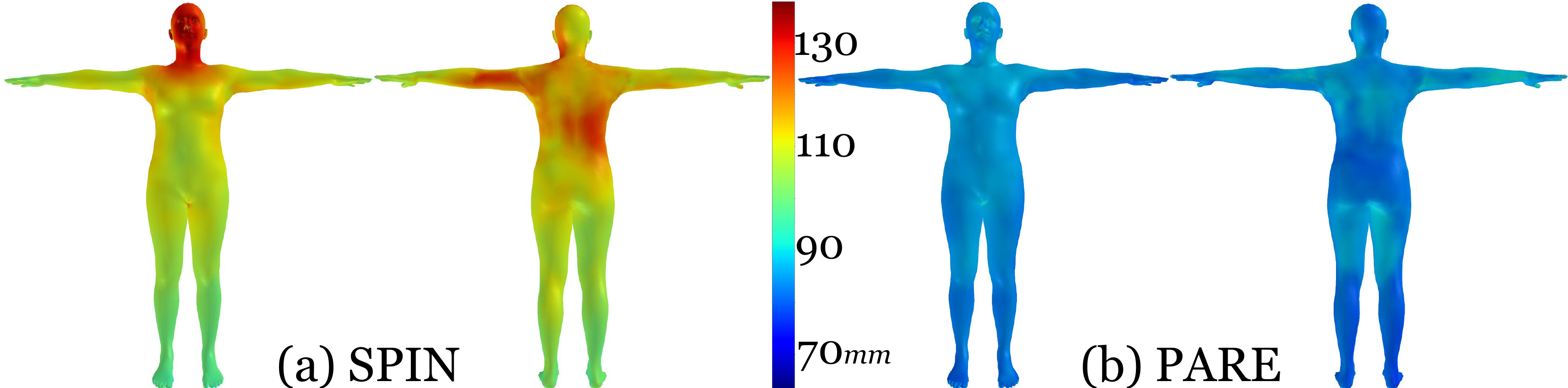}
	\caption{\textbf{Occlusion sensitivity mesh.} Meshes visualize the (a) SPIN and (b) PARE average joint errors.
	}
	\label{fig:occ_meshes_pare}
	\vspace{-2ex}
\end{figure}{}

\begin{figure}[t]
	\centering
	\includegraphics[width=0.9\columnwidth]{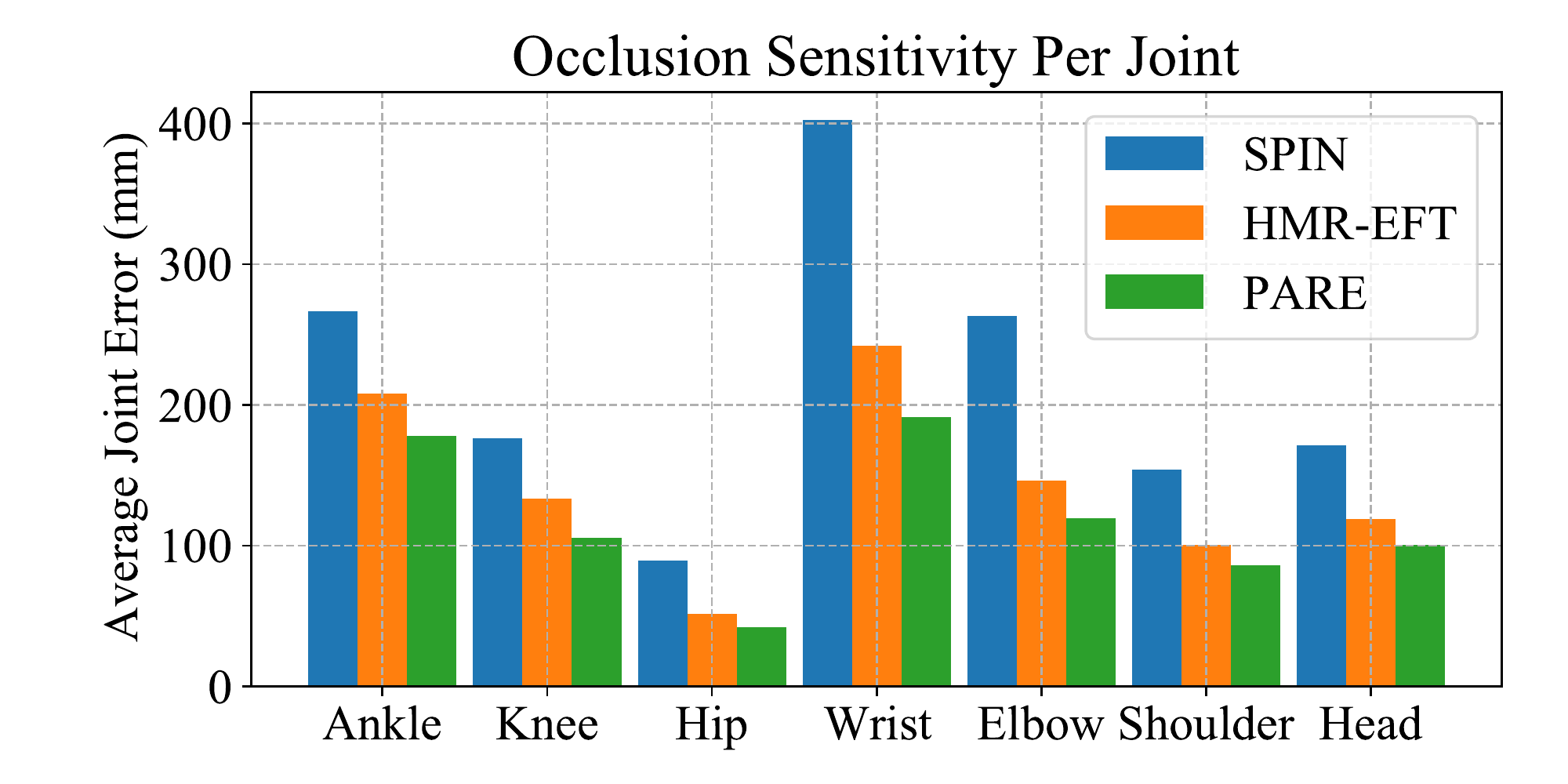}
	\vspace{-0.1in}
	\caption{Per joint occlusion sensitivity analysis of three different methods: SPIN~\cite{SPIN:ICCV:2019}, HMR-EFT~\cite{joo2020eft} (trained with occlusion augmentation), and PARE. PARE is consistently more robust to occlusion.}
	\label{fig:occ_analysis}
	\vspace{-2ex}
\end{figure}{}

\section{Experiments}
\label{experiments}

\paragraph{Training.} 
We train \methodname on \coco~\cite{coco}, \mpii~\cite{mpii}, \lspet~\cite{lspet}, \mpi~\cite{mpiiinf3dhp_mono-2017}, and \hthreesixm~\cite{ionescu_h36m} datasets. More details about these datasets are provided in \supmat Pseudo-ground-truth SMPL annotations for in-the-wild datasets are provided by EFT~\cite{joo2020eft}. 
The part segmentation labels are obtained through rendering segmented SMPL meshes, as visualized in Fig.~\ref{fig:model}. We use 24 parts corresponding to 24 SMPL joints. See \supmat~for samples of part segmentation labels. We used the PyTorch reimplementation~\cite{kolotouros2018pytorch} of Neural Mesh Renderer~\cite{kato2018renderer} to render the parts. For samples without a part segmentation label, we do not supervise the 2D branch.

For the ablation experiments, we train PARE and our baselines on \coco for 175K steps and evaluate on \threedpw and \threedpwocc datasets. We then incorporate all the training data to compare PARE to previous SOTA methods. This pretraining strategy accelerates convergence and reduces the overall training time. It takes about 72 hours to train PARE until convergence on an Nvidia RTX2080Ti GPU. 

To increase robustness to occlusion, we use common occlusion augmentation techniques; \ie~synthetic occlusion (SynthOcc)~\cite{sarandi2018robust} and random crop (RandCrop)~\cite{joo2020eft, Rockwell2020}. All PARE and baseline HMR-EFT models are trained with SynthOcc augmentation unless stated otherwise, \eg~Table~\ref{tab:ablation_occ}.

\noindent
\textbf{Evaluation.} The \threedpw~\cite{vonMarcard2018_3dpw} test split, \threedpwocc~\cite{vonMarcard2018_3dpw, zhangoohcvpr20}, and \ooh~\cite{zhangoohcvpr20} datasets are used for evaluation. We report Procrustes-aligned mean per joint position error (PA-MPJPE) and mean  per  joint position error (MPJPE) in $mm$. For \threedpw we also report per vertex error (PVE) in $mm$.


\begin{table}[t]
	\centering
	\resizebox{0.45\textwidth}{!}{
		\begin{tabular}{ll|r|r|r}
			\toprule
			& & \multicolumn{3}{c}{ \threedpw }  \\
			\cmidrule(lr){3-5}
			& \bf Method & MPJPE $\downarrow$ & PA-MPJPE $\downarrow$ & PVE $\downarrow$\\
			\midrule
			\parbox[t]{2mm}{\multirow{5}{*}{\rotatebox[origin=c]{90}{temporal}}} & HMMR~\cite{kanazawa_hmr} & 116.5 & 72.6 & - \\
			& Doersch~\etal~\cite{doersch_sim2real} & - & 74.7 & - \\
			& Sun~\etal~\cite{Sun_2019_ICCV} & - & 69.5 & - \\
			& VIBE~\cite{kocabas2019vibe} & 93.5 & 56.5 & 113.4 \\
			& MEVA~\cite{Luo2020MEVA} & 86.9 & 54.7 & - \\
			\midrule
			\parbox[t]{2mm}{\multirow{4}{*}{\rotatebox[origin=c]{90}{multi stage}}} & Pose2Mesh~\cite{Choi_2020_ECCV_Pose2Mesh} & 89.2 & 58.9 & - \\
			& Zanfir~\etal~\cite{zanfir2020weakly} & 90.0 & 57.1 & - \\
			& I2L-MeshNet~\cite{Moon_2020_ECCV_I2L-MeshNet} & 93.2 & 58.6  & - \\
			& LearnedGD~\cite{song2020human} & - & 56.4  & - \\
			\midrule
			\parbox[t]{2mm}{\multirow{6}{*}{\rotatebox[origin=c]{90}{single stage}}} & HMR~\cite{kanazawa_hmr} & 130.0 & 76.7  & \\
			& CMR~\cite{kolotouros2019cmr} & - & 70.2  & - \\
			& SPIN~\cite{SPIN:ICCV:2019} & 96.9 & 59.2  & 135.1 \\
			& HMR-EFT~\cite{joo2020eft} & - & 54.2  & - \\
			\cmidrule(lr){2-5}
			& PARE (R50) & 82.9 & 52.3 & 99.7 \\
			& PARE (HRNet-W32) & 82.0 & 50.9 & 97.9 \\
			& PARE (HRNet-W32) w. 3DPW & \textbf{74.5} & \textbf{46.5} & \textbf{88.6} \\
			\bottomrule
		\end{tabular} 
	}
	\vspace{-0.05in}
	\caption{\textbf{Evaluation on the \threedpw dataset.} 
	The units for mean joint and vertex errors are in \em{mm}.
	PARE models outperform temporal, multi-stage, and single-stage state-of-the-art methods. 
	} 
    \label{tab:sota_3dpw}
    \vspace{-2ex}
\end{table}{}

\noindent
\textbf{Comparison to the state-of-the-art.}
Table~\ref{tab:sota_3dpw} compares PARE with previous single-RGB-image HPS estimation methods. We report PARE results with two different backbones: ResNet-50 and HRNet-W32. PARE improves the PA-MPJPE performance by 10\% compared to HMR-EFT~\cite{joo2020eft}, one of the best-performing recent methods.


\begin{table}[t]
	\centering
	\resizebox{\columnwidth}{!}{
		\begin{tabular}{l|r|r|r|r|r}
			\toprule
			& \multicolumn{3}{c|}{ \threedpwocc } & \multicolumn{2}{c}{ \ooh } \\
			\cmidrule(lr){2-6}
			\textbf{Method} & {\footnotesize MPJPE $\downarrow$} & {\footnotesize PA-MPJPE $\downarrow$} & {\footnotesize PVE $\downarrow$} & {\footnotesize MPJPE $\downarrow$} & {\footnotesize PA-MPJPE $\downarrow$} \\
			\midrule
			Zhang~\etal~\cite{zhangoohcvpr20} & - & 72.2 & - & - & 58.5 \\
			SPIN~\cite{SPIN:ICCV:2019} &  95.6 & 60.8 & 121.6 & 104.3 & 68.3 \\
			HMR-EFT~\cite{joo2020eft} & 94.4 & 60.9 & 111.3 & 75.2 & 53.1 \\
			\midrule
			PARE (R50) & \textbf{90.5 }& \textbf{56.6} & \textbf{107.9} & \bf 63.3 & \bf 44.3 \\
			\bottomrule
		\end{tabular} 
	}
	\vspace{-0.05in}
	\caption{\textbf{Evaluation on occlusion datasets \threedpwocc, \ooh}. Here all methods except SPIN are trained with the same datasets, \ie~\coco, \hthreesixm and \ooh. 
	}
	\label{tab:sota_occ}
	\vspace{-2ex}
\end{table}{}


Table~\ref{tab:sota_occ} demonstrates the performance of PARE on occlusion-specific datasets. Here Zhang \etal~\cite{zhangoohcvpr20}, HMR-EFT~\cite{joo2020eft}, and PARE are trained with \coco, \hthreesixm, and \ooh for a fair comparison. We report the SPIN results for reference. HMR-EFT is the fair alternative to SPIN, since SPIN uses HMR as the architecture. PARE consistently improves the performance on these occlusion datasets. 
Although HMR-EFT is trained with exactly the same augmentation and data as PARE, it performs worse.

We also quantify our occlusion sensitivity analysis. Figure \ref{fig:occ_meshes_pare} shows the average joint error of SPIN and PARE methods on the \threedpw test split. SPIN is quite sensitive to upper body occlusions, especially around the head and back. 
PARE is more robust to occlusions and yields lower error overall. See \supmat~for the per-joint version of Fig.~\ref{fig:occ_meshes_pare}. 
Figure \ref{fig:occ_analysis} shows the per-joint breakdown of the mean 3D error from the occlusion sensitivity analysis for three different methods, SPIN, HMR-EFT, and PARE. Here, we retrain HMR-EFT using SynthOcc for a fair comparison. Again, PARE improves the occlusion robustness of all joints.

\begin{figure*}[t]
	\centerline{
		\includegraphics[width=\textwidth]{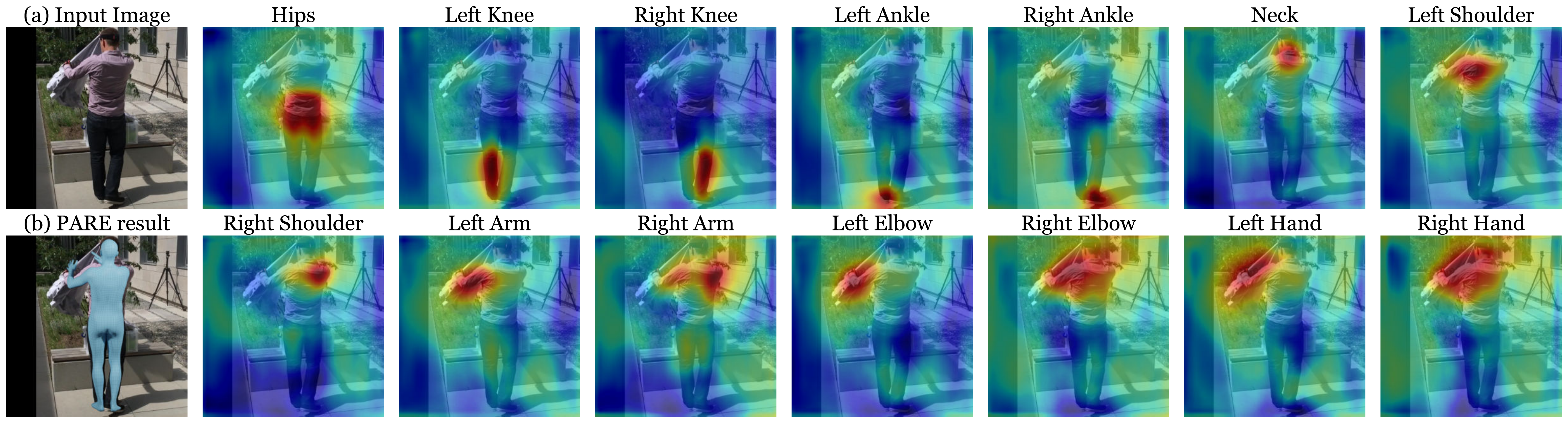}}
	\vspace{-0.1in}
	\caption{\textbf{{\methodname} attention visualization.} 
		Attention maps predicted by the 2D part branch for different joints in image (a).
		For occluded joints like row 2 right hand, PARE learns to attend to larger, more distant, regions to glean information.
	}
	
	\label{fig:attention}
	\vspace{-2ex}
\end{figure*}{}

\begin{figure*}[h]
	\centering
	\includegraphics[width=\textwidth]{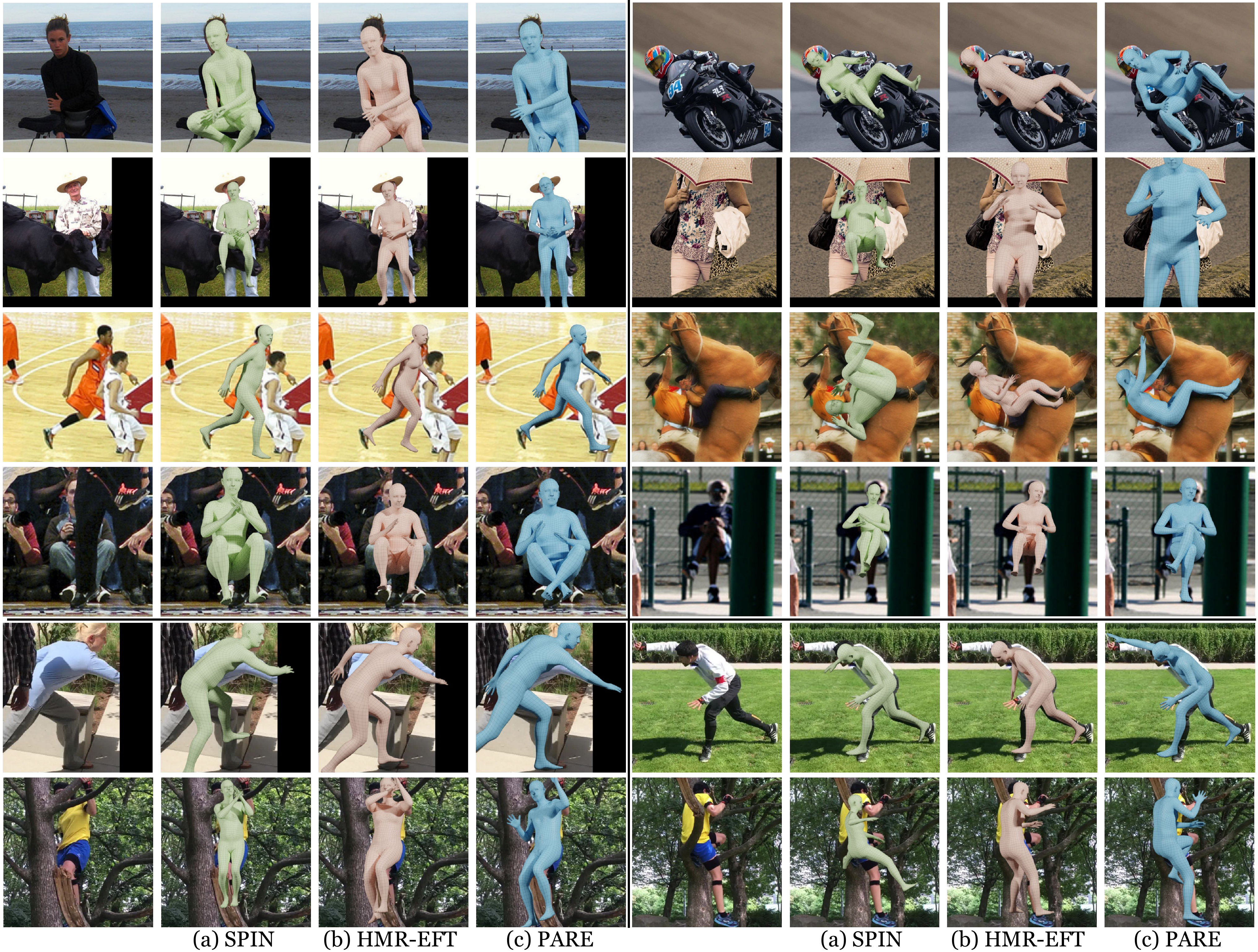}
	\vspace{-0.25in}
	\caption{\textbf{Qualitative results on COCO (rows 1-4) and 3DPW (rows 5-6) datasets.} From left to right: Input image, (a) SPIN~\cite{SPIN:ICCV:2019} results, (b) HMR-EFT~\cite{joo2020eft} results, (c) PARE results.}
	\label{fig:qual}
	\vspace{-1ex}
\end{figure*}

\noindent
\textbf{Qualitative comparison.} We  qualitatively compare SPIN, HMR-EFT, and PARE in Fig.~\ref{fig:qual}.
Even though occlusion augmentation improves robustness to occlusion as seen in the HMR-EFT results, it is not sufficient on its own. PARE, with its attention mechanism, performs well even in challenging occlusion scenarios. More qualitative samples, including failure cases, are provided in \supmat


\begin{table}[t]
	\centering
	\resizebox{\columnwidth}{!}{
	\begin{tabular}{ll|r|r|r|r}
			\toprule 
			& & \multicolumn{2}{c|}{ \threedpw } & \multicolumn{2}{c}{ \threedpwocc } \\
			\cmidrule{3-6}
			\textbf{Method} & & {\footnotesize MPJPE $\downarrow$} & {\footnotesize PA-MPJPE $\downarrow$} & {\footnotesize MPJPE $\downarrow$} & {\footnotesize PA-MPJPE $\downarrow$} \\
			\midrule
			& NBF~\cite{omran2018nbf} & 100.4 & 63.2 & 103.5 & 70.4 \\
			& HMR-EFT & 99.0 & 59.9 & 97.9 & 64.7 \\
			\midrule
			\bf $P$ Supervision & \bf $F$ Sampling & \multicolumn{4}{c}{}  \\
			\midrule
			(a) Joints & Pooling & 95.2 & 58.9 & 95.4 & 63.1 \\
			(b) Joints & Attention & 95.3 & 58.8 & 98.9 & 63.9 \\
			\midrule
			(c) Unsup & Attention & 94.8 & 57.9 & 95.9 & 62.7 \\
			(d) Parts & Attention & 94.5 & 57.3 & 94.7 & \textbf{61.2} \\
			(e) Parts/Unsup & Attention & \textbf{93.4} & \textbf{57.1} & \textbf{93.9} & 61.6 \\
			(f) Parts & Pooling & 97.9 & 59.1 & 99.8 & 64.8 \\
			\bottomrule
		\end{tabular} 
	}
	\vspace{-0.05in}
	\caption{\textbf{Exploring part attention.} 
	The ``$P$ Supervision" column shows the type of supervision for the 2D part branch $P$. ``$F$ Sampling" shows the type of feature sampling method for $F$. All methods are trained on \cocoeft with a ResNet-50 backbone.}
	\vspace{-3ex}
	\label{tab:ablation_attention}

\end{table}{}


\noindent\textbf{Does part attention help?}
Table~\ref{tab:ablation_attention} summarizes our ablation experiments that explore the concept of part attention. First, we compare our results with Neural Body Fitting~\cite{omran2018nbf} trained with identical settings to ours. NBF~\cite{omran2018nbf} can be seen as a straightforward combination of part segmentation and human body regression. Table~\ref{tab:ablation_attention} shows that NBF's two-stage approach is outperformed even by the HMR-EFT baseline. Subsequently, we compare different types of supervision for the 2D part branch $P$ and sampling methods to obtain final features $F^{\prime}$ from $F$. ``Unsup" means $P$ is not supervised.
Inspired by HoloPose~\cite{guler_2019_CVPR}, we first supervise the 2D branch with keypoints and pool the 3D features via bilinear sampling (Table~\ref{tab:ablation_attention}-a). 
Even though this gives lower error than HMR, the improvement is not significant. 
Intuitively, sparse keypoints do not cover enough spatial area to be able to reason about body parts. 
Because the 2D branch predicts Gaussian heatmaps, which cover a larger spatial area than discrete keypoints, we 
explore soft attention instead of pooling to have a larger effective receptive field (Table~\ref{tab:ablation_attention}-b). 
In doing so, however, we do not leverage the full potential of soft attention, which can learn which regions to attend to implicitly from the data. 
So, we remove supervision for the 2D branch to see if soft attention alone can work as well as explicit supervision (Table~\ref{tab:ablation_attention}-c). 
Upon visualizing the resulting attention maps, we find that they are not focused on the body parts.
To induce more structure, we supervise the 2D branch with part segmentation labels (Table~\ref{tab:ablation_attention}-d). 
This approach works significantly better than the above attempts. 
There is a remaining caveat, however: by supervising with a segmentation loss, we constrain the attention map to the parts only, whereas a pure soft attention has the potential to attend to any region it finds informative. 
Consequently, we train with mixed supervision, applying the part segmentation loss for around 125K steps, then continuing to train without supervision (Table~\ref{tab:ablation_attention}-e). 
This final version produces the ``best of both worlds" and the lowest error. We also experiment with part segmentation and pooling to explore the effect of soft-attention (Table~\ref{tab:ablation_attention}-f). Finally, to demonstrate the statistical significance, we performed a two-sided t-test for all experiments in Table~\ref{tab:ablation_attention}; specifically p\textless0.01 for rows (c) vs.~(d), (d) vs.~(e), and (b) vs.~(d).

In addition to joint errors, we measure the mean part segmentation IoU (intersection over union) to better understand how part segmentation and the final pose and shape estimation interact when we do not use part supervision. Mean IoU on the 3DPW test set is 1\%, 85\%, 74\% for (c) unsup, (d) parts, and (e) parts/unsup methods respectively. Lower segmentation accuracy does not hurt the body reconstruction. We provide further body-part segmentation results during different stages of the training in \supmat

Figure \ref{fig:attention} visualizes these attention maps on sample images. 
Part attention learns to attend to body parts or image regions as needed to estimate body shape and pose.


\begin{table}[t]
	\centering
	\resizebox{\columnwidth}{!}{
	    \begin{tabular}{l|r|r|r|r}
	    \toprule
        & \multicolumn{2}{c|}{ \threedpw } & \multicolumn{2}{c}{ \threedpwocc } \\
        \cmidrule{2-5}
        \textbf{Method} & {\footnotesize MPJPE $\downarrow$} & {\footnotesize PA-MPJPE $\downarrow$} & {\footnotesize MPJPE $\downarrow$} & {\footnotesize PA-MPJPE $\downarrow$} \\
        \midrule
        HMR-EFT + SynthOcc &  99.0 & 59.9 & 97.9 & 64.7 \\
        \midrule
        PARE & 95.0 & 57.6 & \textbf{94.4} & 61.3 \\
        PARE + SynthOcc & \textbf{94.5} & \textbf{57.3} & 94.7 & \textbf{61.2} \\
        PARE + SynthOcc + RandCrop & 95.7 & 58.1 & 97.8 & 62.6 \\
        \bottomrule
        \end{tabular} 
	}
	\vspace{-0.05in}
	\caption{\textbf{Ablation of different occlusion augmentation strategies.} We demonstrate the effect of synthetic occlusion (SynthOcc) and random crop (RandCrop) augmentation on the final performance. All methods are trained on \cocoeft with ResNet-50 as the backbone.}
	\label{tab:ablation_occ}
	\vspace{-3ex}
\end{table}{}

\begin{table}[t]
	\centering
	\resizebox{\columnwidth}{!}{
	    \begin{tabular}{l|l|r|r|r|r}
	    \toprule
        \multicolumn{2}{c|}{ } & \multicolumn{2}{c|}{ \threedpw } & \multicolumn{2}{c}{ \threedpwocc } \\
        \cmidrule{3-6}
        \multicolumn{2}{l|}{\textbf{Method}} & {\footnotesize MPJPE $\downarrow$} & {\footnotesize PA-MPJPE $\downarrow$} & {\footnotesize MPJPE $\downarrow$} & {\footnotesize PA-MPJPE $\downarrow$} \\
        \midrule
        HMR-EFT & ResNet-50 & 99.0 & 59.9 & 97.9 & 64.7 \\
        PARE & ResNet-50 & \textbf{93.4} & \textbf{57.1} & \textbf{93.9} & \textbf{61.6} \\
        \midrule
        HMR-EFT & HRNet-W32 & 92.6 & 55.9 & 90.2 & 57.8 \\
        PARE & HRNet-W32 & \textbf{89.0} & \textbf{54.3} & \textbf{87.1} & \textbf{57.0} \\
        \bottomrule
        \end{tabular} 
	}
	\vspace{-0.05in}
	\caption{\textbf{Ablation of backbone architectures}. All methods are trained on \cocoeft.}
	\label{tab:ablation_backbone}
	\vspace{-3ex}
\end{table}{}

{}


\noindent\textbf{Occlusion Augmentation.}
We report the effect of occlusion augmentation techniques in Table~\ref{tab:ablation_occ}. 
SynthOcc improves the performance on both \threedpw and \threedpwocc over vanilla training. Applying RandCrop right at the beginning of the training hurts the performance. Therefore, we start applying crop augmentation after 175K training steps. Between 30\%-50\% of a bounding box is cropped with the probability of 0.3. Even though crop augmentation does not improve performance on \threedpw and \threedpwocc, we find it useful for true in-the-wild images, which often contain significant frame occlusion. 
See \supmat for more examples.


\noindent\textbf{Effect of  CNN backbones.} 
As shown in Table~\ref{tab:ablation_backbone}, HRNet-W32, which produces effective high-resolution representations, performs better than ResNet-50. PARE provides consistent improvements over HMR-EFT with both backbones.

	\vspace{-1.5ex}
\section{Conclusion}
\vspace{-1.5ex}
\label{conclusion}
We present a novel Part Attention Regressor, {\methodname}, which regresses 3D human pose and shape by exploiting information about the visibility  of individual body parts, and thus gaining robustness to occlusion. 
{\methodname} is based on the insights gleaned from our occlusion sensitivity analysis. 
In particular, we observe dependencies between body parts and argue that the network should rely on visible parts to improve predictions for occluded parts and, hence, the overall performance of 3D pose estimation.
Our novel body-part-driven attention mechanism captures such dependencies, using soft attention guided by regressed body part segmentation masks. 
The  network learns to use  part segmentations as visibility  cues  to  reason  about occluded joints and aggregating features from the attended regions. 
This improves  robustness  to  occlusions of different types:  scene, self, and frame occlusion.
Detailed ablation studies show how each choice contributes to our state-of-the-art performance on benchmark datasets. 

	{\small
		\bibliographystyle{ieee_fullname}
		\balance
		\bibliography{main}

\begin{thebibliography}{10}\itemsep=-1pt

\bibitem{agarwal2006recovering}
Ankur Agarwal and Bill Triggs.
\newblock Recovering {3D} human pose from monocular images.
\newblock {\em {IEEE} Transaction on Pattern Analysis and Machine
  Intelligence}, 28(1):44--58, 2006.

\bibitem{mpii}
Mykhaylo Andriluka, Leonid Pishchulin, Peter Gehler, and Bernt Schiele.
\newblock 2d human pose estimation: New benchmark and state of the art
  analysis.
\newblock In {\em IEEE Conference on Computer Vision and Pattern Recognition},
  2014.

\bibitem{scape}
Dragomir Anguelov, Praveen Srinivasan, Daphne Koller, Sebastian Thrun, Jim
  Rodgers, and James Davis.
\newblock Scape: Shape completion and animation of people.
\newblock {\em SIGGRAPH}, 2005.

\bibitem{balan2008}
Alexandru Balan and Michael~J Black.
\newblock The naked truth: Estimating body shape under clothing.
\newblock In {\em European Conference on Computer Vision}, 2008.

\bibitem{Balan:CVPR:2007}
Alexandru~O Balan, Leonid Sigal, Michael~J Black, James~E Davis, and Horst~W
  Haussecker.
\newblock Detailed human shape and pose from images.
\newblock In {\em IEEE Conference on Computer Vision and Pattern Recognition}.
  IEEE, 2007.

\bibitem{biggs2020multibodies}
Benjamin Biggs, David Novotny, Sebastien Ehrhardt, Hanbyul Joo, Ben Graham, and
  Andrea Vedaldi.
\newblock 3d multi-bodies: Fitting sets of plausible 3d human models to
  ambiguous image data.
\newblock In {\em Advances in Neural Information Processing}, 2020.

\bibitem{bogo_smplify}
Federica Bogo, Angjoo Kanazawa, Christoph Lassner, Peter Gehler, Javier Romero,
  and Michael~J. Black.
\newblock Keep it {SMPL}: Automatic estimation of {3D} human pose and shape
  from a single image.
\newblock In {\em European Conference on Computer Vision}, 2016.

\bibitem{cheng20203d}
Yu Cheng, Bo Yang, Bo Wang, and Robby~T Tan.
\newblock 3d human pose estimation using spatio-temporal networks with explicit
  occlusion training.
\newblock {\em arXiv preprint arXiv:2004.11822}, 2020.

\bibitem{cheng2019occlusion}
Yu Cheng, Bo Yang, Bo Wang, Wending Yan, and Robby~T Tan.
\newblock Occlusion-aware networks for 3d human pose estimation in video.
\newblock In {\em International Conference on Computer Vision}, pages 723--732,
  2019.

\bibitem{Choi_2020_ECCV_Pose2Mesh}
Hongsuk Choi, Gyeongsik Moon, and Kyoung~Mu Lee.
\newblock Pose2{M}esh: Graph convolutional network for 3d human pose and mesh
  recovery from a 2d human pose.
\newblock In {\em European Conference on Computer Vision}, pages 769--787,
  2020.

\bibitem{expose2020eccv}
Vasileios Choutas, Georgios Pavlakos, Timo Bolkart, Dimitrios Tzionas, and
  Michael~J. Black.
\newblock Monocular expressive body regression through body-driven attention.
\newblock In {\em European Conference on Computer Vision}, pages 20--40, 2020.

\bibitem{doersch_sim2real}
Carl Doersch and Andrew Zisserman.
\newblock Sim2real transfer learning for {3D} pose estimation: motion to the
  rescue.
\newblock In {\em Advances in Neural Information Processing}, 2019.

\bibitem{georgakis2020hierarchical}
Georgios Georgakis, Ren Li, Srikrishna Karanam, Terrence Chen, Jana Kosecka,
  and Ziyan Wu.
\newblock Hierarchical kinematic human mesh recovery.
\newblock In {\em European Conference on Computer Vision}, 2020.

\bibitem{ghiasi2014parsing}
Golnaz Ghiasi, Yi Yang, Deva Ramanan, and Charless~C Fowlkes.
\newblock Parsing occluded people.
\newblock In {\em IEEE Conference on Computer Vision and Pattern Recognition},
  2014.

\bibitem{grauman2003inferring}
Kristen Grauman, Gregory Shakhnarovich, and Trevor Darrell.
\newblock Inferring {3D} structure with a statistical image-based shape model.
\newblock In {\em International Conference on Computer Vision}, pages 641--648,
  2003.

\bibitem{guler_2019_CVPR}
Riza~Alp Guler and Iasonas Kokkinos.
\newblock {HoloPose}: Holistic {3D} human reconstruction in-the-wild.
\newblock In {\em IEEE Conference on Computer Vision and Pattern Recognition},
  pages 10884--10894, 2019.

\bibitem{he2016resnet}
Kaiming He, Xiangyu Zhang, Shaoqing Ren, and Jian Sun.
\newblock Identity mappings in deep residual networks.
\newblock In {\em European Conference on Computer Vision}, 2016.

\bibitem{epipolartransformers2020cvpr}
Yihui He, Rui Yan, Katerina Fragkiadaki, and Shoou-I Yu.
\newblock Epipolar transformers.
\newblock In {\em IEEE Conference on Computer Vision and Pattern Recognition},
  pages 7779--7788, 2020.

\bibitem{huang2009estimating}
Jia-Bin Huang and Ming-Hsuan Yang.
\newblock Estimating human pose from occluded images.
\newblock In {\em Asian Conference on Computer Vision}, pages 48--60, 2009.

\bibitem{ionescu_h36m}
Catalin Ionescu, Dragos Papava, Vlad Olaru, and Cristian Sminchisescu.
\newblock {Human3.6M}: Large scale datasets and predictive methods for {3D}
  human sensing in natural environments.
\newblock In {\em {IEEE} Transaction on Pattern Analysis and Machine
  Intelligence}, 2014.

\bibitem{jiang2020mpshape}
Wen Jiang, Nikos Kolotouros, Georgios Pavlakos, Xiaowei Zhou, and Kostas
  Daniilidis.
\newblock Coherent reconstruction of multiple humans from a single image.
\newblock In {\em IEEE Conference on Computer Vision and Pattern Recognition},
  2020.

\bibitem{lspet}
Sam Johnson and Mark Everingham.
\newblock Learning effective human pose estimation from inaccurate annotation.
\newblock In {\em IEEE Conference on Computer Vision and Pattern Recognition},
  2011.

\bibitem{joo2020eft}
Hanbyul Joo, Natalia Neverova, and Andrea Vedaldi.
\newblock Exemplar fine-tuning for {3D} human pose fitting towards in-the-wild
  {3D} human pose estimation.
\newblock {\em {arXiv}:2004.03686}, 2020.

\bibitem{kanazawa_hmr}
Angjoo Kanazawa, Michael~J. Black, David~W. Jacobs, and Jitendra Malik.
\newblock End-to-end recovery of human shape and pose.
\newblock In {\em IEEE Conference on Computer Vision and Pattern Recognition},
  pages 7122--7131, 2018.

\bibitem{kanazawa_temporal_hmr}
Angjoo Kanazawa, Jason~Y. Zhang, Panna Felsen, and Jitendra Malik.
\newblock Learning {3D} human dynamics from video.
\newblock In {\em IEEE Conference on Computer Vision and Pattern Recognition},
  pages 5614--5623, 2019.

\bibitem{kato2018renderer}
Hiroharu Kato, Yoshitaka Ushiku, and Tatsuya Harada.
\newblock Neural 3d mesh renderer.
\newblock In {\em IEEE Conference on Computer Vision and Pattern Recognition},
  2018.

\bibitem{kocabas2019vibe}
Muhammed Kocabas, Nikos Athanasiou, and Michael~J. Black.
\newblock {VIBE}: Video inference for human body pose and shape estimation.
\newblock In {\em IEEE Conference on Computer Vision and Pattern Recognition},
  pages 5252--5262, 2020.

\bibitem{kolotouros2018pytorch}
Nikos Kolotouros.
\newblock Pytorch implementation of the neural mesh renderer, 2018.

\bibitem{SPIN:ICCV:2019}
Nikos Kolotouros, Georgios Pavlakos, Michael~J. Black, and Kostas Daniilidis.
\newblock Learning to reconstruct {3D} human pose and shape via model-fitting
  in the loop.
\newblock In {\em International Conference on Computer Vision}, pages
  2252--2261, 2019.

\bibitem{kolotouros2019cmr}
Nikos Kolotouros, Georgios Pavlakos, and Kostas Daniilidis.
\newblock Convolutional mesh regression for single-image human shape
  reconstruction.
\newblock In {\em IEEE Conference on Computer Vision and Pattern Recognition},
  pages 4501--4510, 2019.

\bibitem{lassner_up3d}
Christoph Lassner, Javier Romero, Martin Kiefel, Federica Bogo, Michael~J.
  Black, and Peter~V. Gehler.
\newblock Unite the {People}: Closing the loop between {3D} and {2D} human
  representations.
\newblock In {\em IEEE Conference on Computer Vision and Pattern Recognition},
  pages 4704--4713, 2017.

\bibitem{coco}
Tsung-Yi Lin, Michael Maire, Serge Belongie, James Hays, Pietro Perona, Deva
  Ramanan, Piotr Doll{\'a}r, and C~Lawrence Zitnick.
\newblock Microsoft {COCO}: Common objects in context.
\newblock In {\em European Conference on Computer Vision}, 2014.

\bibitem{looper_smpl}
Matthew Loper, Naureen Mahmood, Javier Romero, Gerard Pons-Moll, and Michael~J.
  Black.
\newblock {SMPL}: A skinned multi-person linear model.
\newblock In {\em ACM Trans. Graphics (Proc. SIGGRAPH Asia)}, 2015.

\bibitem{Lu_2019_CVPR}
Xiankai Lu, Wenguan Wang, Chao Ma, Jianbing Shen, Ling Shao, and Fatih Porikli.
\newblock See more, know more: Unsupervised video object segmentation with
  co-attention siamese networks.
\newblock In {\em IEEE Conference on Computer Vision and Pattern Recognition},
  2019.

\bibitem{Luo2020MEVA}
Zhengyi Luo, S. Golestaneh, and Kris~M. Kitani.
\newblock {3D} human motion estimation via motion compression and refinement.
\newblock In {\em Asian Conference on Computer Vision}, pages 324--340, 2020.

\bibitem{mahendran2015understanding}
Aravindh Mahendran and Andrea Vedaldi.
\newblock Understanding deep image representations by inverting them.
\newblock In {\em IEEE Conference on Computer Vision and Pattern Recognition},
  pages 5188--5196, 2015.

\bibitem{mpiiinf3dhp_mono-2017}
Dushyant Mehta, Helge Rhodin, Dan Casas, Pascal Fua, Oleksandr Sotnychenko,
  Weipeng Xu, and Christian Theobalt.
\newblock Monocular {3D} human pose estimation in the wild using improved {CNN}
  supervision.
\newblock In {\em International Conference on 3DVision}, 2017.

\bibitem{Moon_2020_ECCV_I2L-MeshNet}
Gyeongsik Moon and Kyoung~Mu Lee.
\newblock {I2L-MeshNet}: Image-to-lixel prediction network for accurate 3d
  human pose and mesh estimation from a single rgb image.
\newblock In {\em European Conference on Computer Vision}, pages 752--768,
  2020.

\bibitem{omran2018nbf}
Mohamed Omran, Christoph Lassner, Gerard Pons-Moll, Peter~V. Gehler, and Bernt
  Schiele.
\newblock Neural body fitting: Unifying deep learning and model-based human
  pose and shape estimation.
\newblock In {\em International Conference on 3DVision}, 2018.

\bibitem{SMPL-X:2019}
Georgios Pavlakos, Vasileios Choutas, Nima Ghorbani, Timo Bolkart, Ahmed A.~A.
  Osman, Dimitrios Tzionas, and Michael~J. Black.
\newblock Expressive body capture: {3D} hands, face, and body from a single
  image.
\newblock In {\em IEEE Conference on Computer Vision and Pattern Recognition},
  pages 10975--10985, 2019.

\bibitem{pavlakos2018humanshape}
Georgios Pavlakos, Luyang Zhu, Xiaowei Zhou, and Kostas Daniilidis.
\newblock Learning to estimate 3{D} human pose and shape from a single color
  image.
\newblock In {\em IEEE Conference on Computer Vision and Pattern Recognition},
  pages 459--468, 2018.

\bibitem{Leonid2016DeepCut}
Leonid Pishchulin, Eldar Insafutdinov, Siyu Tang, Bjoern Andres, Mykhaylo
  Andriluka, Peter Gehler, and Bernt Schiele.
\newblock {DeepCut}: Joint subset partition and labeling for multi person pose
  estimation.
\newblock In {\em IEEE Conference on Computer Vision and Pattern Recognition},
  2016.

\bibitem{rafi2015semantic}
Umer Rafi, Juergen Gall, and Bastian Leibe.
\newblock A semantic occlusion model for human pose estimation from a single
  depth image.
\newblock In {\em Proceedings of the IEEE Conference on Computer Vision and
  Pattern Recognition Workshops}, 2015.

\bibitem{Rockwell2020}
Chris Rockwell and David~F. Fouhey.
\newblock Full-body awareness from partial observations.
\newblock In {\em European Conference on Computer Vision}, pages 522--539,
  2020.

\bibitem{sarandi2018robust}
Istv{\'a}n S{\'a}r{\'a}ndi, Timm Linder, Kai~O Arras, and Bastian Leibe.
\newblock How robust is 3d human pose estimation to occlusion?
\newblock {\em IROS workshops}, 2018.

\bibitem{selvaraju2017grad}
Ramprasaath~R Selvaraju, Michael Cogswell, Abhishek Das, Ramakrishna Vedantam,
  Devi Parikh, and Dhruv Batra.
\newblock Grad-cam: Visual explanations from deep networks via gradient-based
  localization.
\newblock In {\em International Conference on Computer Vision}, 2017.

\bibitem{sengupta2020synthetic}
Akash Sengupta, Ignas Budvytis, and Roberto Cipolla.
\newblock Synthetic training for accurate 3d human pose and shape estimation in
  the wild.
\newblock {\em British Machine Vision Conference}, 2020.

\bibitem{sigal2008combined}
Leonid Sigal, Alexandru Balan, and Michael~J Black.
\newblock Combined discriminative and generative articulated pose and non-rigid
  shape estimation.
\newblock In {\em Advances in Neural Information Processing}, 2008.

\bibitem{song2020human}
Jie Song, Xu Chen, and Otmar Hilliges.
\newblock Human body model fitting by learned gradient descent.
\newblock In {\em European Conference on Computer Vision}, 2020.

\bibitem{hrnet}
Ke Sun, Bin Xiao, Dong Liu, and Jingdong Wang.
\newblock Deep high-resolution representation learning for human pose
  estimation.
\newblock In {\em IEEE Conference on Computer Vision and Pattern Recognition},
  2019.

\bibitem{Sun_2019_ICCV}
Yu Sun, Yun Ye, Wu Liu, Wenpeng Gao, Yili Fu, and Tao Mei.
\newblock Human mesh recovery from monocular images via a skeleton-disentangled
  representation.
\newblock In {\em International Conference on Computer Vision}, 2019.

\bibitem{Tan}
Jun Kai~Vince Tan, Ignas Budvytis, and Roberto Cipolla.
\newblock Indirect deep structured learning for {3D} human shape and pose
  prediction.
\newblock In {\em British Machine Vision Conference}, 2017.

\bibitem{tung2017self}
Hsiao-Yu Tung, Hsiao-Wei Tung, Ersin Yumer, and Katerina Fragkiadaki.
\newblock Self-supervised learning of motion capture.
\newblock In {\em Advances in Neural Information Processing}, pages 5236--5246,
  2017.

\bibitem{vonMarcard2018_3dpw}
Timo von Marcard, Roberto Henschel, Michael Black, Bodo Rosenhahn, and Gerard
  Pons-Moll.
\newblock Recovering accurate 3d human pose in the wild using imus and a moving
  camera.
\newblock In {\em European Conference on Computer Vision}, pages 614--631,
  2018.

\bibitem{vosoughi2018deep}
Saeid Vosoughi and Maria~A Amer.
\newblock Deep 3d human pose estimation under partial body presence.
\newblock In {\em 2018 25th IEEE International Conference on Image Processing
  (ICIP)}, pages 569--573. IEEE, 2018.

\bibitem{wang20203d}
Justin Wang, Edward Xu, Kangrui Xue, and Lukasz Kidzinski.
\newblock 3{D} pose detection in videos: Focusing on occlusion.
\newblock {\em arXiv preprint arXiv:2006.13517}, 2020.

\bibitem{Wang_2018_CVPR}
Xiaolong Wang, Ross Girshick, Abhinav Gupta, and Kaiming He.
\newblock Non-local neural networks.
\newblock In {\em IEEE Conference on Computer Vision and Pattern Recognition},
  pages 7794--7803, 2018.

\bibitem{CenterHMR}
Sun Yu, Bao Qian, Liu Wu, Fu Yili, and Mei Tao.
\newblock Center{HMR}: a bottom-up single-shot method for multi-person 3d mesh
  recovery from a single image.
\newblock 2020.

\bibitem{zanfir2020weakly}
Andrei Zanfir, Eduard~Gabriel Bazavan, Hongyi Xu, Bill Freeman, Rahul
  Sukthankar, and Cristian Sminchisescu.
\newblock Weakly supervised 3d human pose and shape reconstruction with
  normalizing flows.
\newblock In {\em European Conference on Computer Vision}, pages 465--481,
  2020.

\bibitem{zeiler2014visualizing}
Matthew~D Zeiler and Rob Fergus.
\newblock Visualizing and understanding convolutional networks.
\newblock In {\em European Conference on Computer Vision}, 2014.

\bibitem{zhangoohcvpr20}
Tianshu Zhang, Buzhen Huang, and Yangang Wang.
\newblock Object-occluded human shape and pose estimation from a single color
  image.
\newblock In {\em IEEE Conference on Computer Vision and Pattern Recognition},
  pages 7374--7383, 2020.

\bibitem{zhou2016learning}
Bolei Zhou, Aditya Khosla, Agata Lapedriza, Aude Oliva, and Antonio Torralba.
\newblock Learning deep features for discriminative localization.
\newblock In {\em IEEE Conference on Computer Vision and Pattern Recognition},
  pages 2921--2929, 2016.

\end{thebibliography}
	}
	\newpage
	\appendix
	{\noindent\Large\textbf{Supplementary Material}}
	\newline
	\setcounter{page}{1}
	
The Supplementary Material consists of this document and a video. They include acknowledgement, disclosure, additional information and visualizations of our method and results.

\noindent
{\bf Acknowledgements:}
We thank Joachim Tesch for helping with Blender rendering. We thank Nikos Athanasiou, Vassilis Choutas, Emre Aksan, Stefan Stevsics, Xu Chen, Cornelia Kohler, Marilyn Keller, Shashank Tripathi, Yinghao Huang, Omid Taheri, Sai Kumar Dwivedi, Hongwei Yi, Dimitris Tzionas, Timo Bolkart, Yao Feng, and all Perceiving Systems department members for their feedback and fruitful discussions. This research was partially supported by the Max Planck ETH Center for Learning Systems.

\noindent{\bf Disclosure:} \url{https://files.is.tue.mpg.de/black/CoI/ICCV2021.txt}
\section{Methods}
\paragraph{Implementation Details.}
In all our experiments, we use the weights pretrained on \mpii~\cite{mpii} for a 2D pose estimation task to initialize both ResNet-50 and HRNet-W32, because we observe slower convergence with ImageNet pretrained weights. 
For Table 3-5 in the main paper, we train PARE and our baselines on \coco for 175K steps and evaluate on \threedpw and \threedpwocc datasets. We then include all the training data for the SOTA experiment in Table 1 of the main paper. 
For Table 2, we use the training data of \cite{zhangoohcvpr20} to align the experiment settings.

\paragraph{Loss.}
We use different weight coefficients $\lambda$ for each term in the loss function. They are $\lambda_{3D}=300$, $\lambda_{2D}=300$, $\lambda_{SMPL}=60$, $\lambda_P=60$.

\begin{figure}[t]
	\centering
	\includegraphics[width=0.5\textwidth]{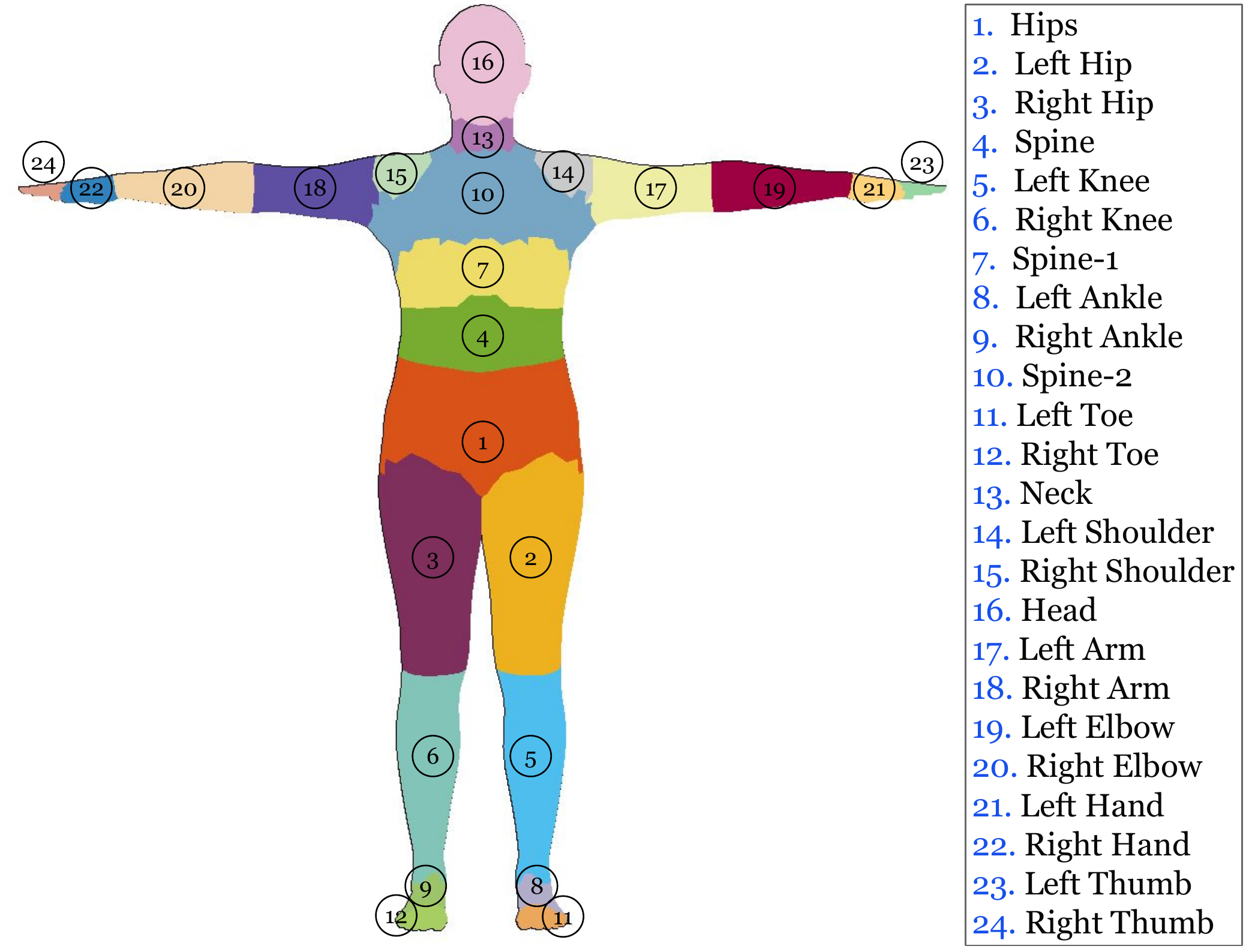}
	\caption{Body part segmentation labels used for the 2D part branch. For each joint in the SMPL kinematic tree, we have a body part label. Correspondences between joints \textit{(right)} and body part labels \textit{(left)} are shown in this figure.}
	\label{fig:segmentation}
\end{figure}

\paragraph{Body Part Segmentation labels.}
Since we have SMPL annotations for most of the samples in our datasets, we do not need additional body part segmentation annotations. We directly use the SMPL annotations to obtain supervision. In Fig~\ref{fig:segmentation}, we visualize this body part labels. For each joint in the SMPL kinematic tree, we have a corresponding body part label.

{}

\begin{figure*}[t]
	\centering
	\includegraphics[width=\textwidth]{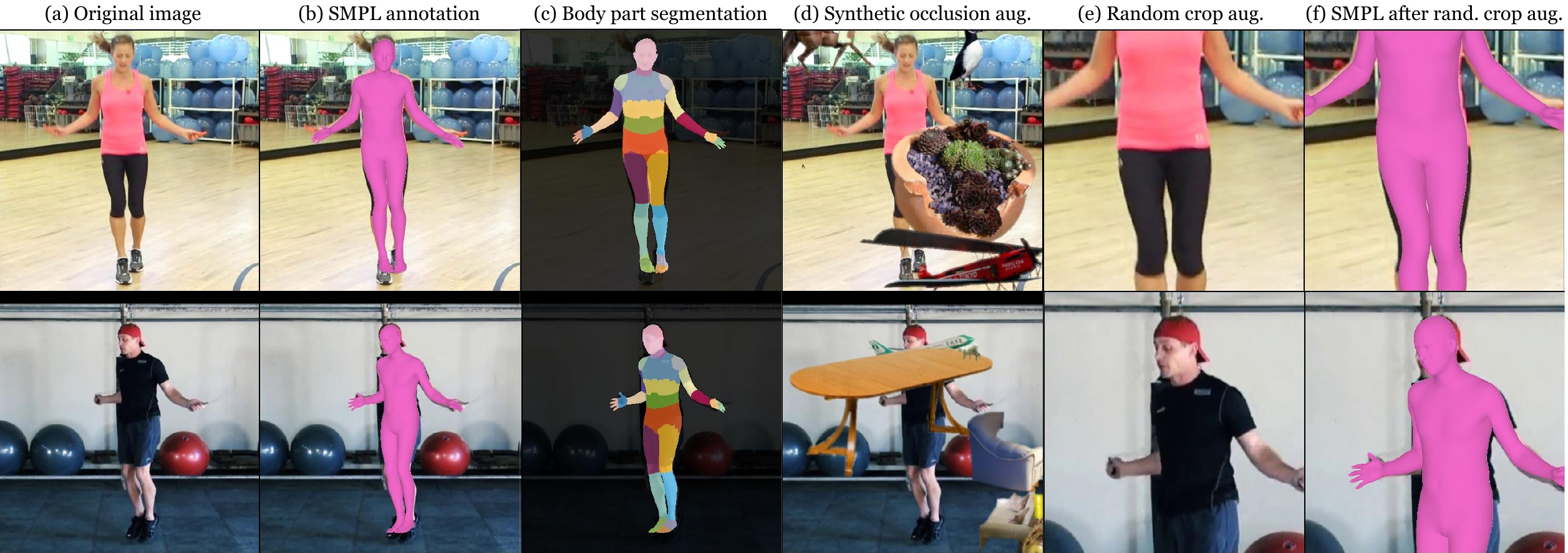}
	\caption{Training samples after synthetic occlusion and random crop augmentations are applied.}
	\label{fig:occ_aug}
\end{figure*}

\paragraph{Occlusion augmentation.}
In Fig.~\ref{fig:occ_aug}, we demonstrate the results of synthetic occlusion and random crop augmentations on two sample images. 
{}

\paragraph{Runtime} PARE is only 1 ms/image slower than HMR, with runtime of 14.8ms on a GTX2080Ti. 

\section{Experiments}
\subsection{Training Datasets}
Our training datasets closely follow previous work, namely EFT~\cite{joo2020eft}, SPIN~\cite{SPIN:ICCV:2019}, and HMR~\cite{kanazawa_hmr}. Here we provide the details for completeness.

\noindent\textbf{\mpi}~\cite{mpiiinf3dhp_mono-2017} is a multi-view indoor 3D human pose estimation dataset. 3D annotations are captured via a commercial markerless mocap software, therefore it is less accurate than some of the 3D datasets \eg \hthreesixm~\cite{ionescu_h36m}. We use all of the training subjects S1 to S8 which makes 90K images in total.

\noindent\textbf{\hthreesixm}~\cite{ionescu_h36m} is an indoor, multi-view 3D human pose estimation dataset. Following previous methods, for training, we use 5 subjects (S1, S5, S6, S7, S8) which means 292K images.

\noindent\textbf{In-the-wild 2D datasets} \coco~\cite{coco}, \mpii~\cite{mpii} and \lspet~\cite{lspet} are in-the-wild 2D keypoint datasets. \mpii has 14K, \coco has 75K, \lspet has 7K instances labeled with 2D keypoints. In addition to 2D keypoint annotations, we utilize the pseudo SMPL annotations provided by the EFT~\cite{joo2020eft} method.

\paragraph{Training Dataset Ratios.}
To obtain the final best performing model, we follow EFT~\cite{joo2020eft} and SPIN~\cite{SPIN:ICCV:2019} which use fixed data sampling ratios for each batch. After training 100\% with \cocoeft for 175K steps, we incorporate 50\% \hthreesixm, 30\% In-the-wild (\ie [\coco, \mpii, \lspet]-EFT), and 20\% \mpi datasets into training. We also observe that using [50\% \hthreesixm, 30\% \cocoeft, 20\% \mpi] or [20\% \hthreesixm, 30\% \cocoeft, 50\% \mpi] gives equivalent performance on the \threedpw dataset.

\begin{figure*}[t]
	\centering
	\includegraphics[width=\textwidth]{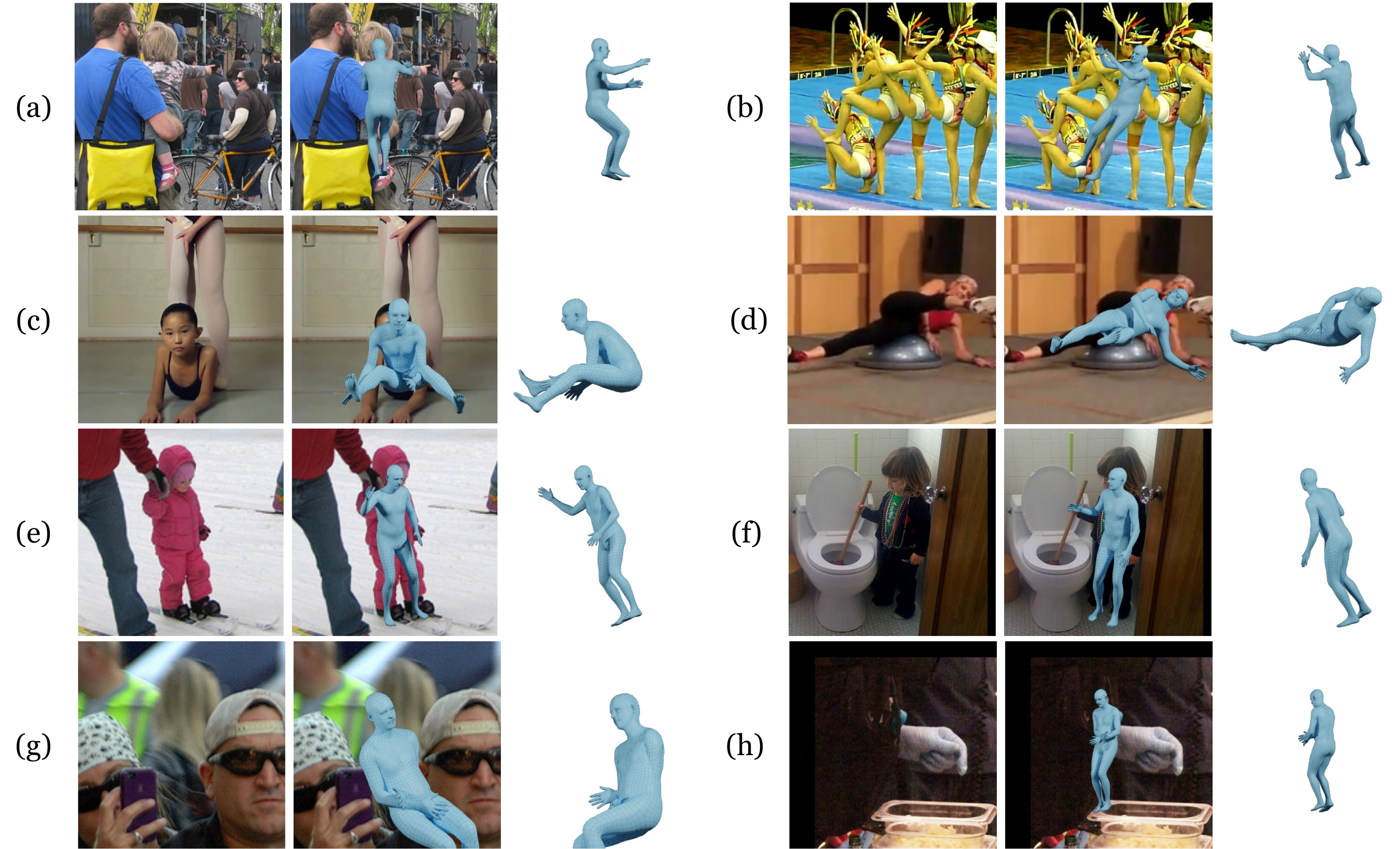}
	\caption{Challenging scenarios where \methodname fails to produce fairly good reconstructions.}
	\label{fig:failure}
\end{figure*}

\paragraph{Failure Cases.}
In Fig.~\ref{fig:failure}, we show a few examples where \methodname fails to reconstruct reasonable human body poses. The scenarios range from (a-b) too many people in the crop, (b-d) rarely-seen extreme poses, (e-f) children whose body shapes cannot be fully explained by the SMPL model, and (g-h) extreme occlusion.

\begin{figure*}
	\centering
	\includegraphics[width=\textwidth]{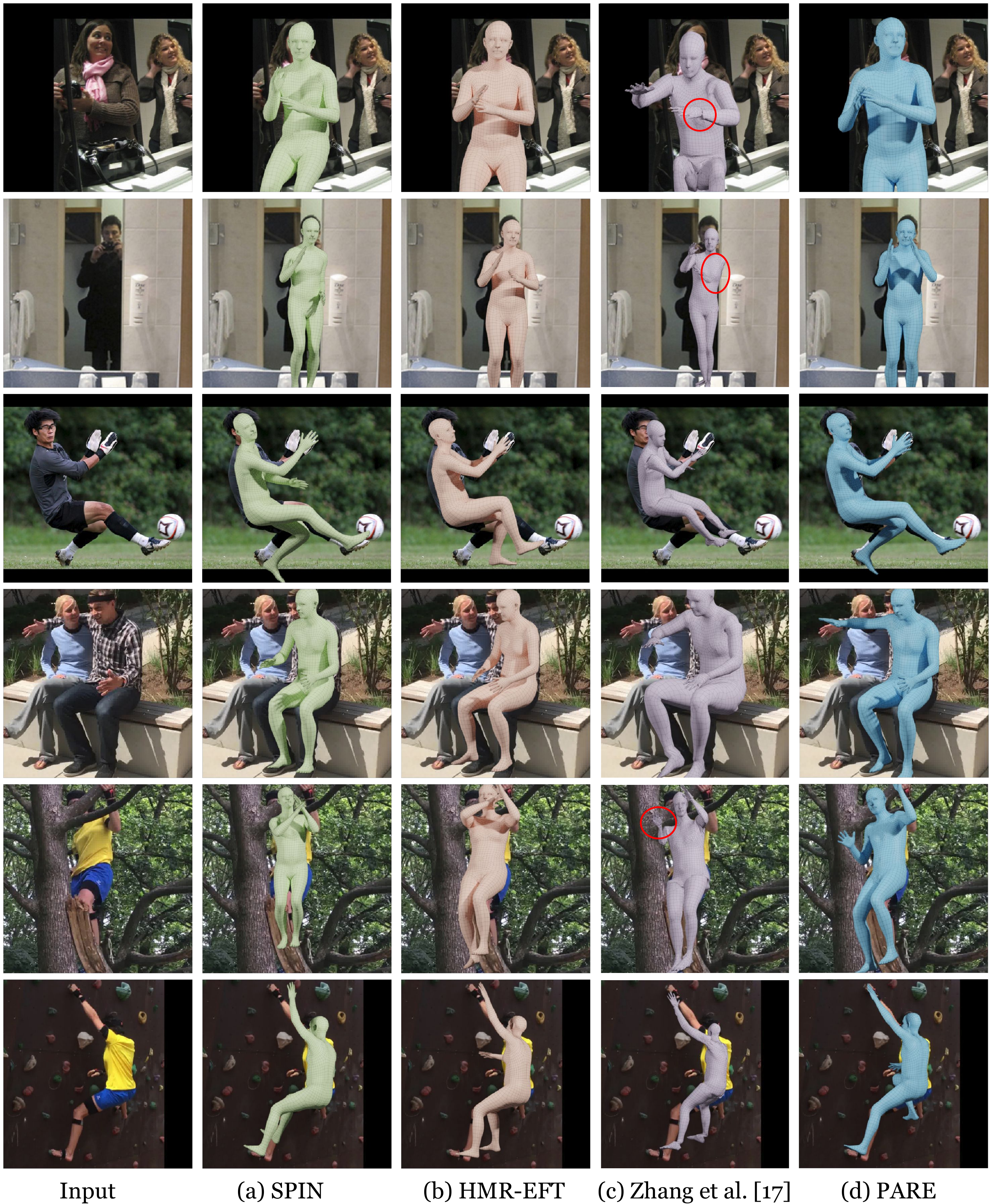}
	\caption{\textbf{Qualitative comparison.} Here, we compare \methodname with recent state-of-the-art methods \ie SPIN~\cite{SPIN:ICCV:2019}, HMR-EFT~\cite{joo2020eft}, and Zhang \etal~\cite{zhangoohcvpr20}.}
	\label{fig:quali-results}
\end{figure*}

\paragraph{Comparing to \cite{zhangoohcvpr20}.}
Zhang \etal~\cite{zhangoohcvpr20} parameterize human meshes as UV maps where each pixel stores the 3D location of a vertex.
They leverage saliency masks as visibility information and cast occlusions as an image-inpainting problem. 
However, we find that the raw predicted vertex locations from \cite{zhangoohcvpr20} yield arbitrary global scale, rotation, translation, and there are no camera parameters associated with the output. 
What they show in the main paper are the post-processed results after fitting a SMPL model, which is not described in their released implementation;  how to visualize the unprocessed, raw predicted meshes is unclear.
Thus, we bring them to the same camera coordinate frame as \methodname through Procrustes Alignment and overlay them on the input as shown in Fig.~\ref{fig:quali-results}(c).
One clearly sees mesh artifacts (red ovals), which is common for non-parametric models. 
The requirement of accurate saliency maps certainly limits the performance of \cite{zhangoohcvpr20} on in-the-wild images.

\paragraph{Comparing to state-of-the-art Temporal Models.}
\begin{table}[h]
	\centering
	\resizebox{0.42\textwidth}{!}{
		\begin{tabular}{ll|r|r}
			\toprule
			& & \multicolumn{2}{c}{ \threedpw }  \\
			\cmidrule(lr){3-4}
			& \bf Method & MPJPE $\downarrow$ & PA-MPJPE $\downarrow$ \\
			\midrule
			\parbox[t]{2mm}{\multirow{5}{*}{\rotatebox[origin=c]{90}{Temporal}}} & HMMR~\cite{kanazawa_hmr} & 116.5 & 72.6  \\
			& Doersch~\etal~\cite{doersch_sim2real} & - & 74.7 \\
			& Sun~\etal~\cite{Sun_2019_ICCV} & - & 69.5 \\
			& VIBE~\cite{kocabas2019vibe} & 93.5 & 56.5  \\
			& MEVA~\cite{Luo2020MEVA} & 86.9 & 54.7 \\
			\midrule
			& PARE (ResNet-50) & 82.9 & 52.3  \\
			& PARE (HRNet-W32) & \bf 82.0 & \bf 50.9  \\
			\bottomrule
		\end{tabular} 
	}
	\caption{\textbf{Evaluation on the \threedpw dataset.} The numbers are average joint errors in mm. PARE models outperform video-based methods which leverage temporal information.}
    \label{tab:sota_video}
    \vspace{-2ex}
\end{table}{}

In Table~\ref{tab:sota_video}, we compare PARE to recent state-of-the-art video based models. 
To do so, we run a SOTA multi-object tracker and then run PARE independently on each frame of the tracklets, with no  temporal smoothing.
Even the SOTA video methods have access to extra temporal information, PARE outperforms them. 
We show some qualitative results of VIBE and PARE on some challenging images in Fig~\ref{fig:vibe}.
Please see the supplemental video for a better visualization of the video results (starts at \texttt{05:21}).
{}

\begin{table}[h]
	\centering
	\resizebox{0.3\textwidth}{!}{
		\begin{tabular}{l|r|r|r}
			\toprule
			& SPIN~\cite{SPIN:ICCV:2019} & HMR-EFT & PARE \\
			\midrule
			PCK $\uparrow$ & 81.5 & 83.4 & \textbf{85.1}  \\
			\bottomrule
		\end{tabular} 
	}
	\caption{\textbf{Evaluation of 2D keypoint project accuracy on \threedpw dataset.}}
	\label{tab:keypoint_2d}
	\vspace{-2ex}
\end{table}{}

\paragraph{2D keypoint projection accuracy} 
We evaluate the 2D keypoint accuracy of our method by projecting the 3D keypoints to the image space using the estimated camera parameters on 3DPW test set. Percentage of correct keypoints (PCK) is used as the evaluation metric. The results are reported in Table~\ref{tab:keypoint_2d}.

\section{More on visualizing attention of networks}
Two new visualizations are proposed in this work: (1) an occlusion sensitivity map/mesh and (2) a part attention map. We provide more examples and discussions for both visualizations.
Please see the video for an animation of the sensitivity analysis, which more clearly illustrates the approach.

\paragraph{Occlusion Sensitivity.}
There are many visualization techniques \cite{mahendran2015understanding,selvaraju2017grad,zeiler2014visualizing,zhou2016learning} available to inspect what CNNs learn. We are, however, more interested in studying how perturbations in the input image affect the output rather than visualizing the internal filters learned by CNNs. We therefore follow the framework of \cite{zeiler2014visualizing} and replace the classification score with an error measure for body poses, as described in the main paper. 
We choose MPJPE as the error measure \emph{without} Procrustes Alignment, because PA-MPJPE leads to artificially low error by aligning global orientations, which are a major source of error.

\begin{figure*}[t]
	\centering
	\includegraphics[width=\textwidth]{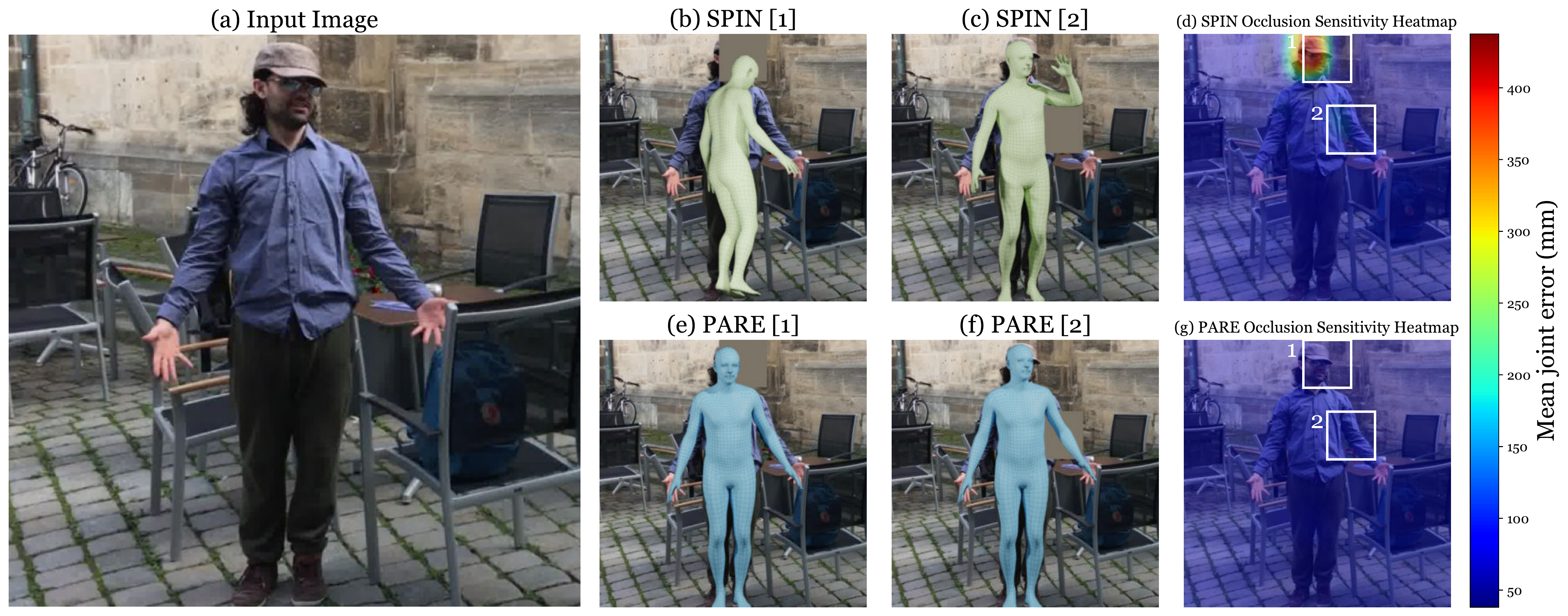}
	\includegraphics[width=\textwidth]{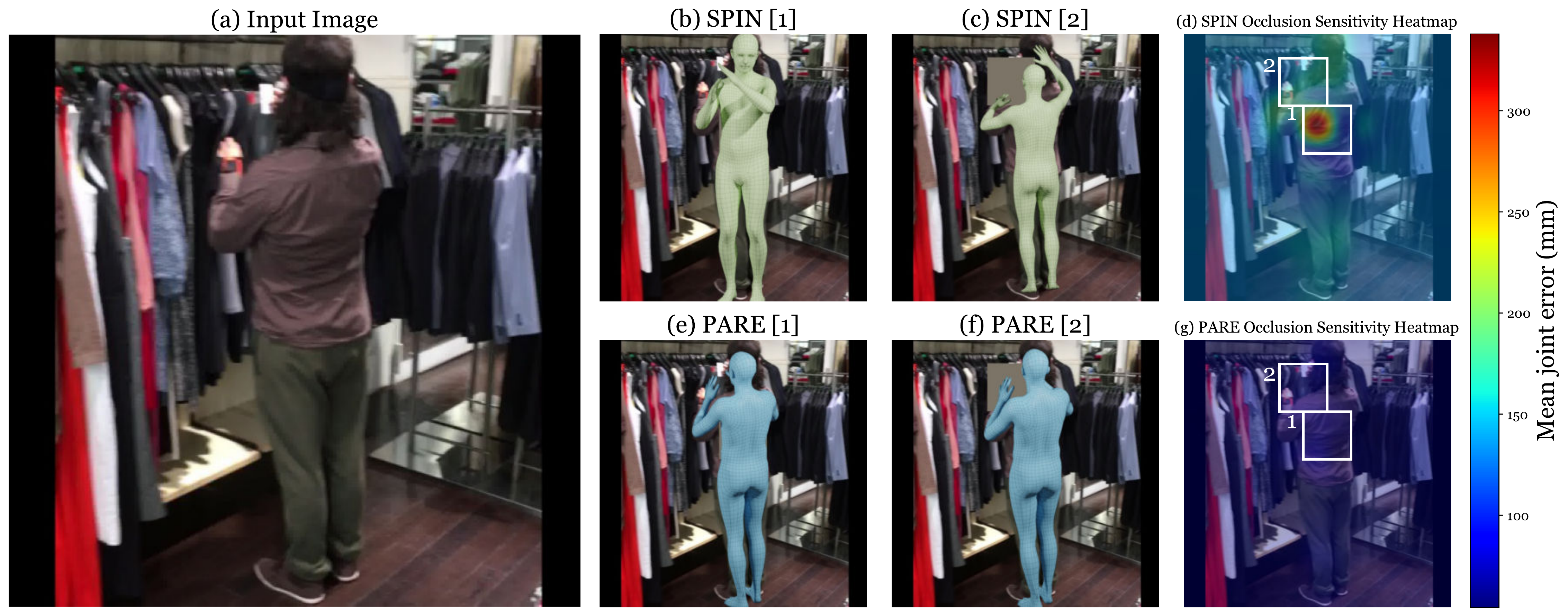}
	\includegraphics[width=\textwidth]{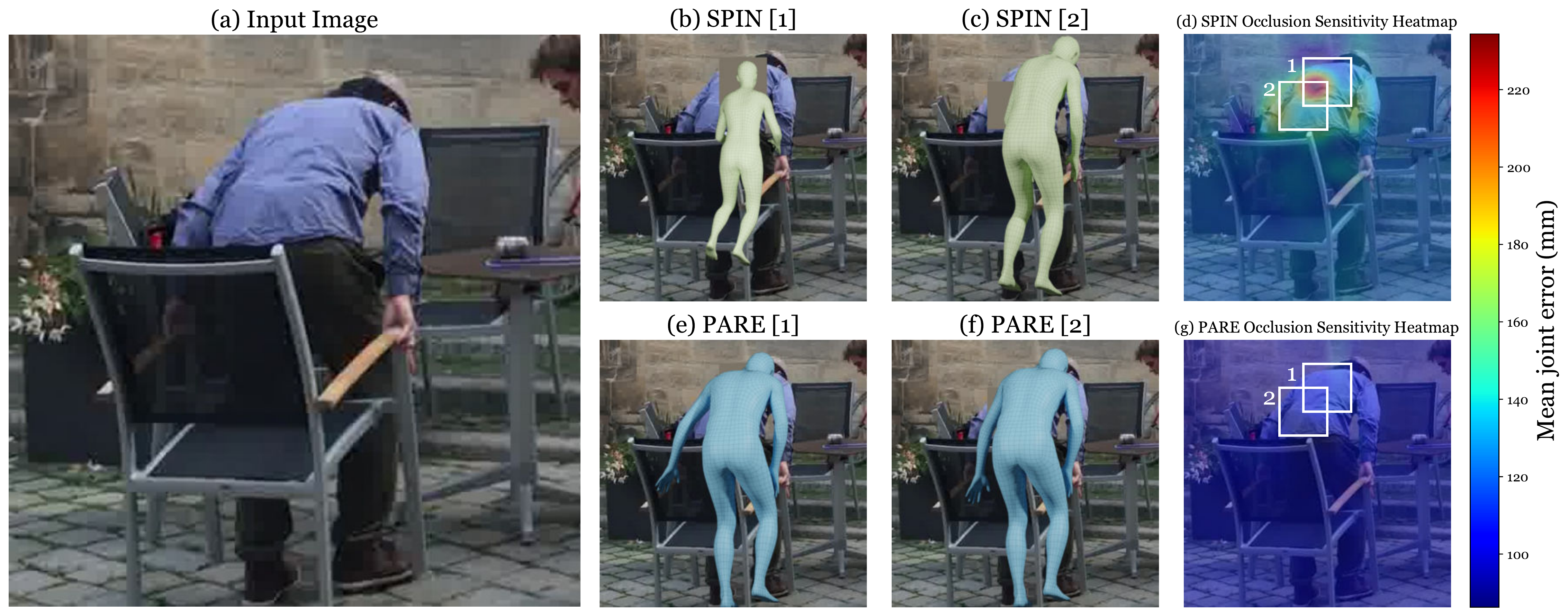}
	\caption{Occlusion Sensitivity Maps of SPIN \cite{SPIN:ICCV:2019} and \methodname}
	\label{fig:teaser-like}
\end{figure*}

This analysis is not limited to a particular network architecture so we also apply it on \methodname and visualize the error maps together with those from SPIN \cite{SPIN:ICCV:2019} in Fig.~\ref{fig:teaser-like}.
Warmer colors correspond to higher MPJPEs w.r.t. ground truth when those pixels are occluded, suggesting that methods rely on the regions to estimate body poses. 
One clearly sees that \methodname is more robust to localized part occlusion. Please see the video for animation (starts at \texttt{00:53}).

\begin{figure*}[t]
	\centering
	\includegraphics[width=0.9\textwidth]{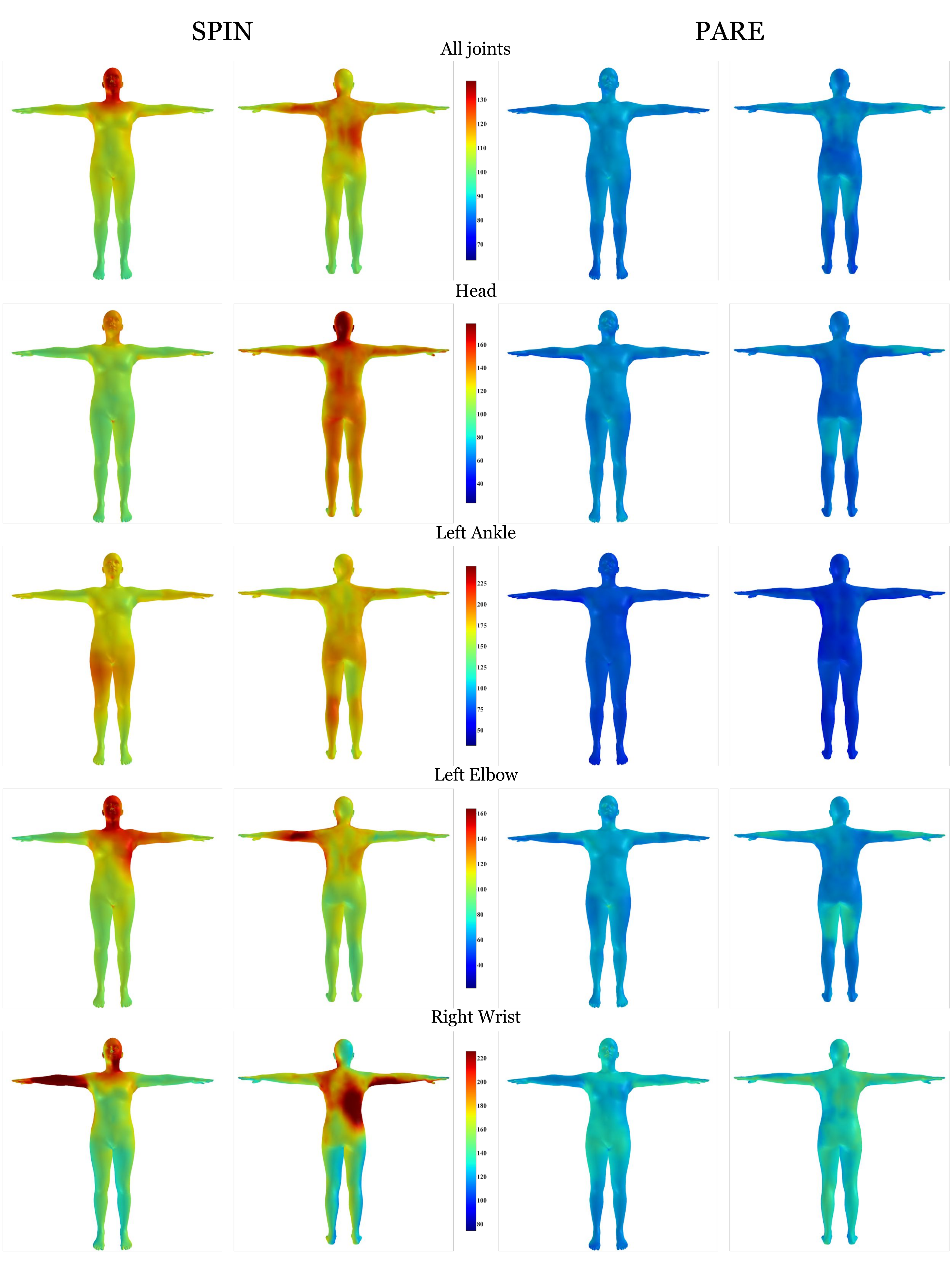}
	\caption{Occlusion sensitivity meshes per joint.}
	\label{fig:occ_mesh_per_joint}
\end{figure*}

Additionally, we also map the per-pixel error to the overlaying 3D vertex, and aggregate over the whole \threedpw dataset \cite{vonMarcard2018_3dpw}. 
In this way, we visualize the per-joint error on the SMPL template mesh, which we term the \emph{occlusion sensitivity mesh}.
Fig.~\ref{fig:occ_mesh_per_joint} shows the occlusion sensitivity mesh for four different joints and averaged over all joints from both SPIN and \methodname. 
We again observe that SPIN is very sensitive to localized part occlusion. For example, occlusions of right arm or face regions result in high error for right wrist. 
On the other hand, occlusion sensitivity meshes of \methodname have more consistent cold colors over the body, again confirming that it is more robust to partial occlusion.

\begin{figure}[t]
	\centering
	\includegraphics[width=\columnwidth]{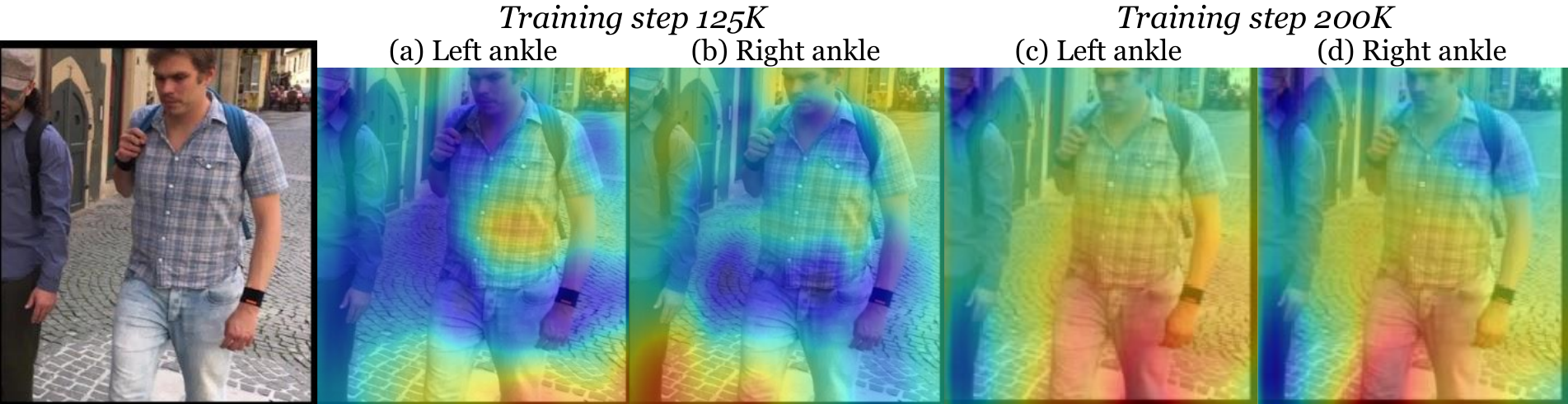}
	\caption{Attention map progression during training. Training uses body-part supervision only until step 125K (a-b). Note that the final attention maps  for occluded parts (at 200K (c-d)) focus on visible parents.} 
	\label{fig:attention_rebuttal}
\end{figure}

\begin{figure*}[h]
	\centering
	\includegraphics[width=\textwidth]{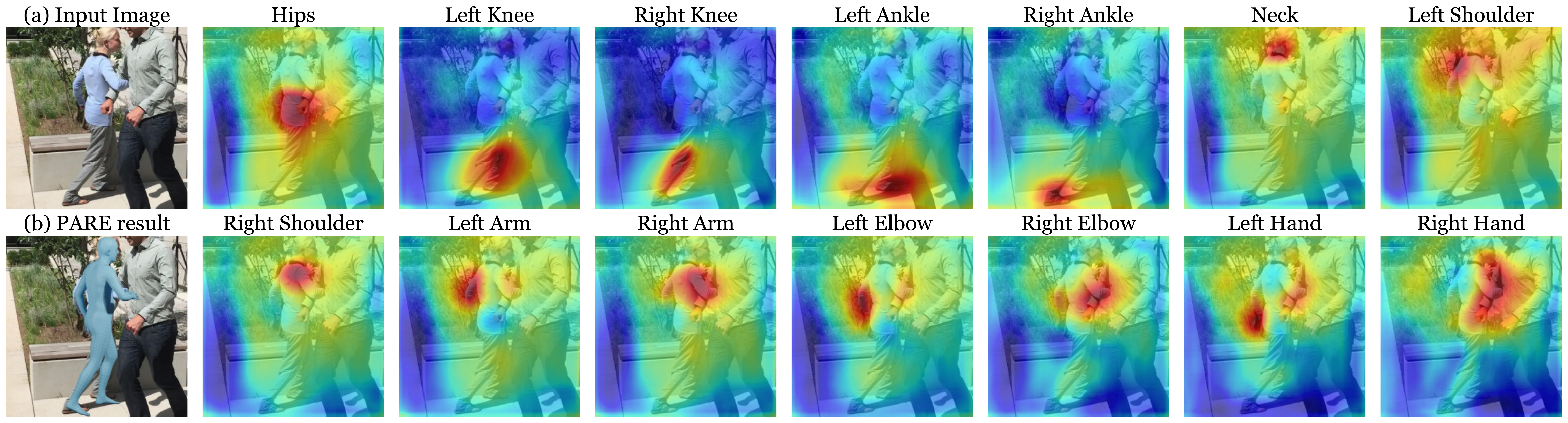}
	\includegraphics[width=\textwidth]{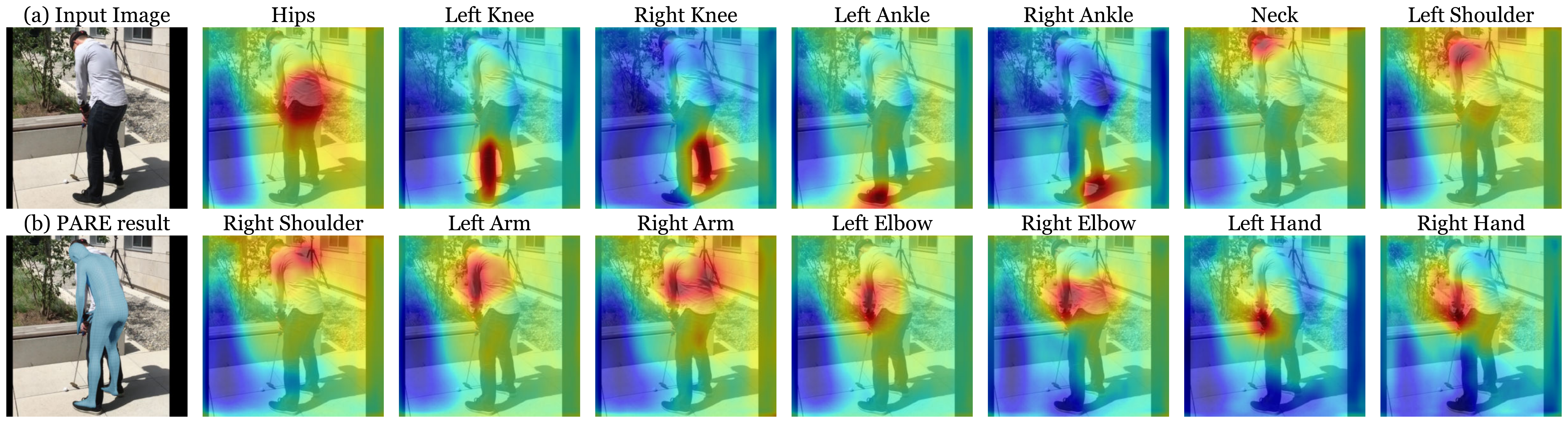}
	\includegraphics[width=\textwidth]{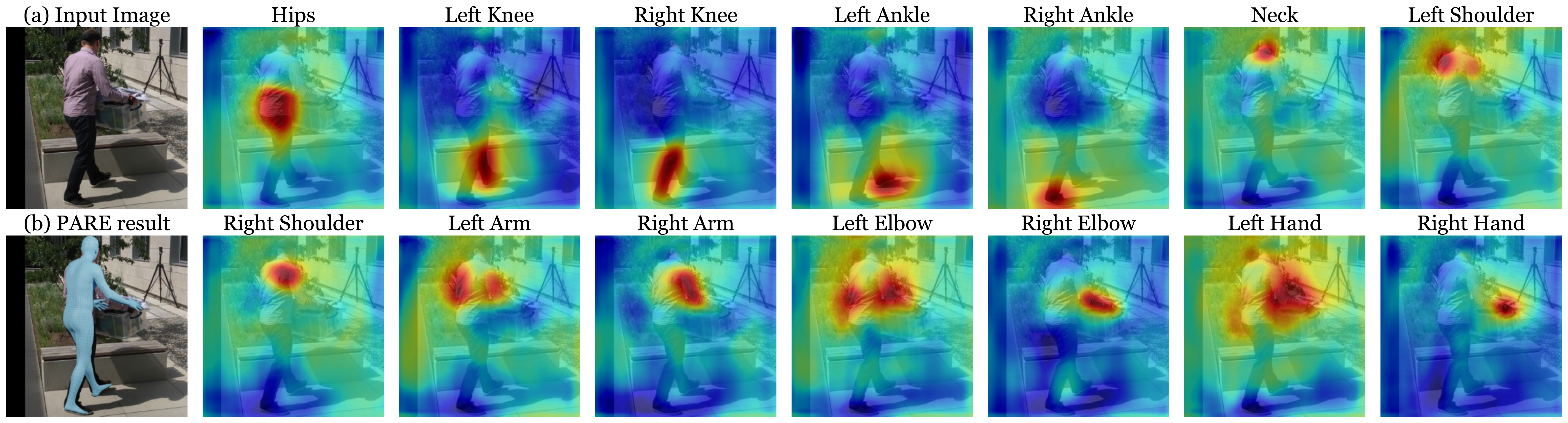}
	\includegraphics[width=\textwidth]{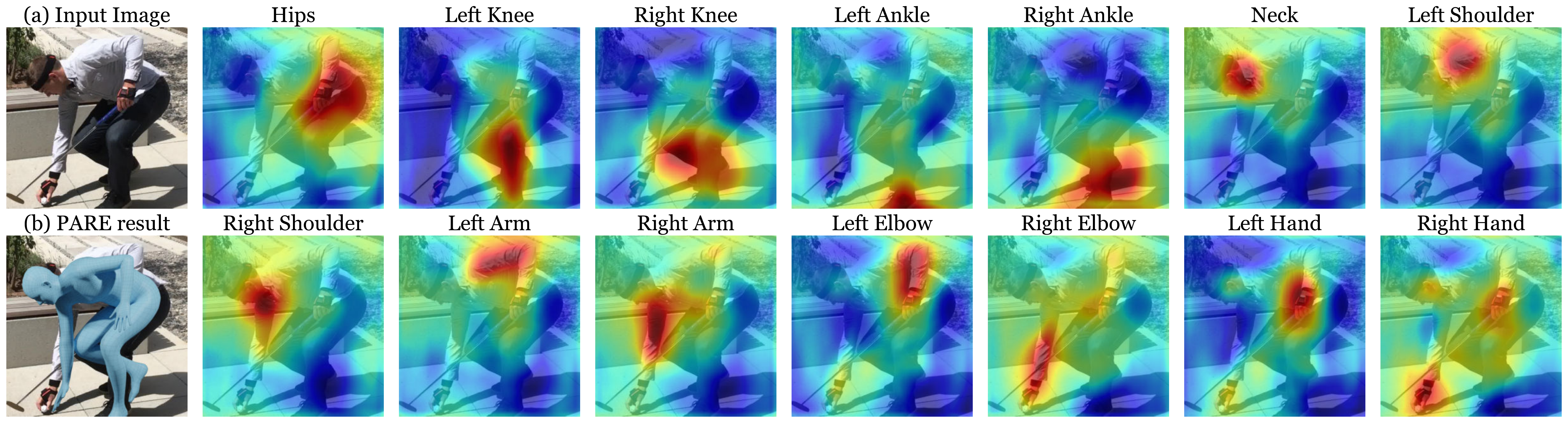}
	\caption{Part attention maps.}
	\label{fig:attention_map}
\end{figure*}

\paragraph{Part Attention.}
We also visualize the estimated part attention $P$ before softmax in Fig.~\ref{fig:attention_map} for four sample images from \threedpw \cite{vonMarcard2018_3dpw}. 
When body parts are visible, the shapes of warm regions resemble part segmentation labels, which means the network focuses on body part regions (\eg~Left/Right Knee and Ankle in the third row). For naturally occluded body parts, the attended regions get wider, covering other parts and the scene. 
This suggests that \methodname implicitly learns to reason about the visibility of body parts and leverages available information to solve the task.
In particular, Fig.~\ref{fig:attention_rebuttal} illustrates the progression of attention maps during training for two occluded parts Left/Right Ankles.
We see that deactivating the part supervision helps attention maps to focus on more meaningful and explainable regions.

\begin{figure*}[h]
	\centering
	\includegraphics[width=\textwidth]{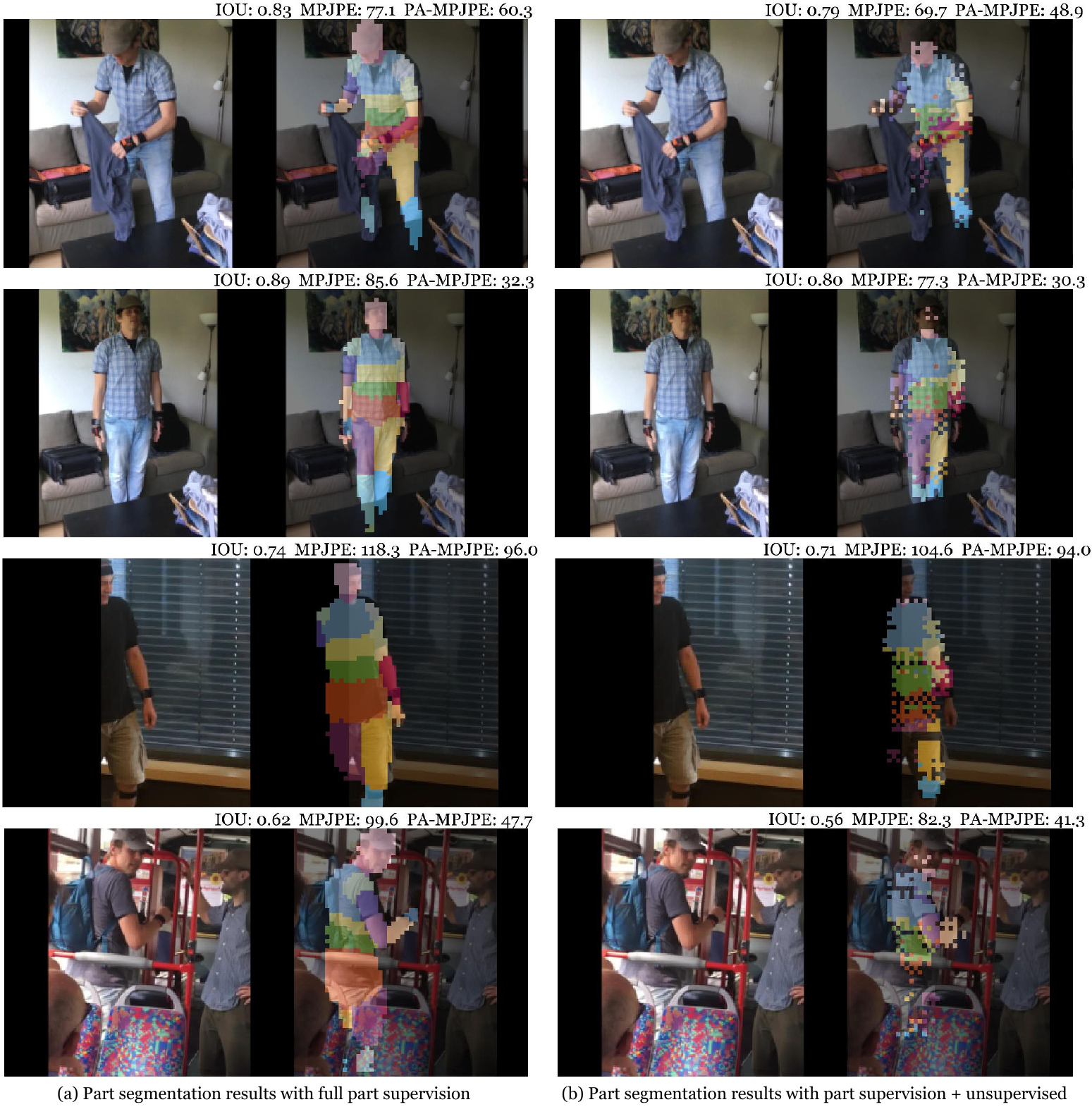}
	\caption{Part segmentation results in two different scenarios: (a) full part segmentation supervision is applied during training, (b) part segmentation supervision is applied at the initial stages and training is continued without part supervision. At the top of each result, we denote the part segmentation IoU, MPJPE and PA-MPJPE. Notice how part segmentation IoU decreases, but per-joint accuracy improves.}
	\label{fig:segmentation_map}
\end{figure*}

In addition to part attention maps, we also visualize the results as segmentation maps in Fig.~\ref{fig:segmentation_map}. We visualize the results of two different models; (a) a model trained with full part segmentation supervision, (b) a model trained with part segmentation initially and unsupervised for the final stages. Note that part segmentation IoU decreases significantly when we do not use part segmentation, however we see an increase in body reconstruction accuracy especially in the case of occlusion.
{}

\newpage
{}

\begin{figure*}[t]
	\centering
	\includegraphics[width=\textwidth]{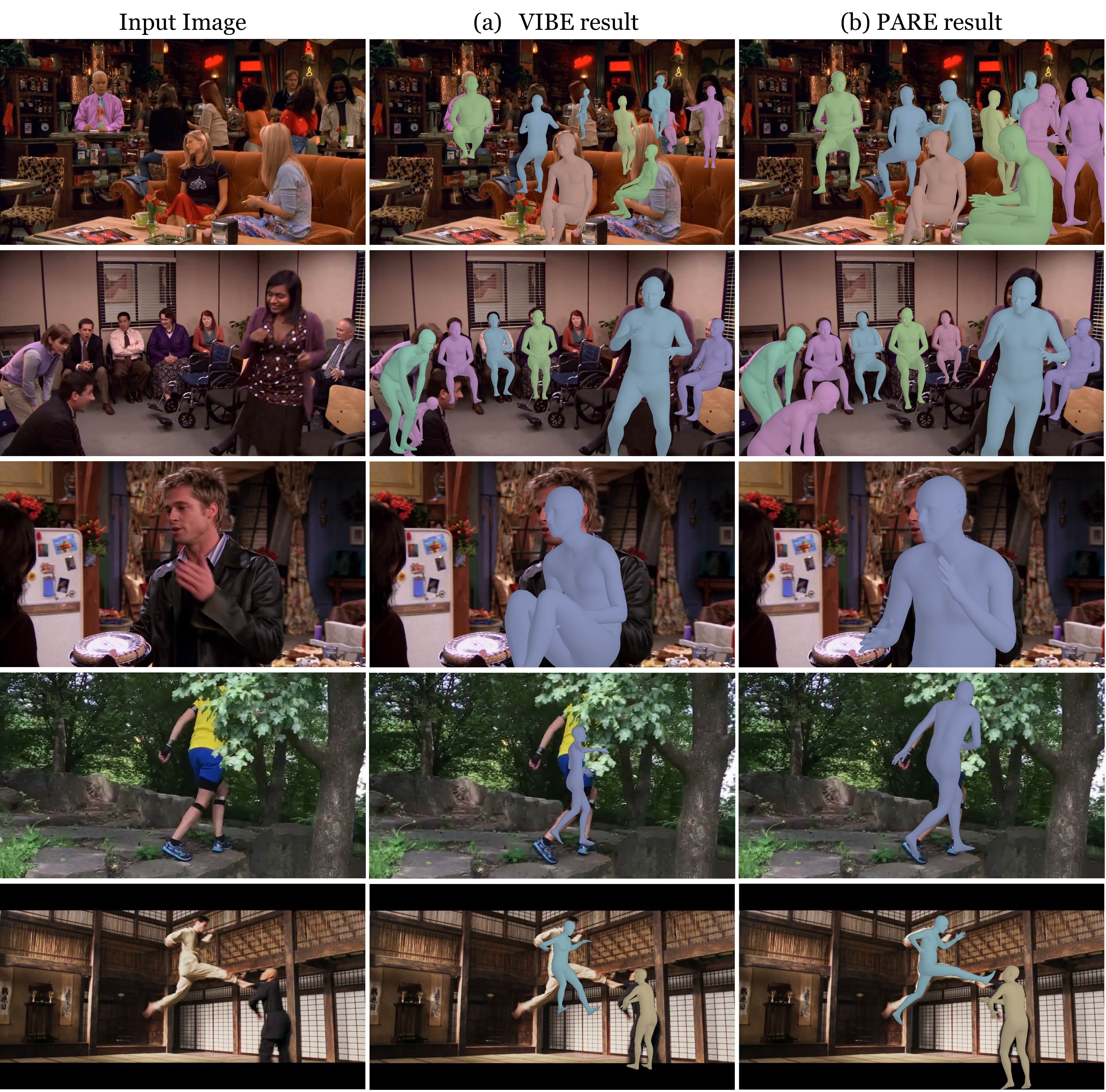}
	\caption{Comparison of VIBE~\cite{kocabas2019vibe} with our method, \methodname. Note that VIBE is a video-based method, while PARE is run on each video frame independently.}
	\label{fig:vibe}
\end{figure*}{}

{}

{}

{}

{}
	
\end{document}